\documentclass[10pt, conference]{IEEEtran}
\usepackage[utf8]{inputenc} 
\usepackage[T1]{fontenc}  
\usepackage{mathrsfs}
\usepackage{%
algorithm,%
algpseudocode,%
amsfonts,%
amsmath,%
amssymb,%
amsthm,%
bbm,%
caption,%
cite,%
comment,%
extarrows,%
float,%
graphicx,%
indentfirst,%
listings,%
paralist,%
pgfplots,%
subcaption,%
tikz,%
todonotes,%
url,%
xcolor,%
xspace,
makecell
}

\usepackage{epstopdf}

\pgfplotsset{plot coordinates/math parser=false}
\usepgflibrary{%
patterns,%
}

\usetikzlibrary{%
arrows,%
automata,%
chains,%
decorations.pathreplacing,%
intersections,%
positioning,%
shapes,%
}

\IEEEoverridecommandlockouts  

\newtheorem{lemma}{Lemma}
\newtheorem{theorem}{Theorem}
\newtheorem{cor}{Corollary}
\newtheorem{proposition}{Proposition}
\newtheorem{assumption}{Assumption}

\theoremstyle{definition}
\newtheorem{definition}{Definition}

\theoremstyle{remark}
\newtheorem{remark}{Remark}

\definecolor{darkgreen}{rgb}{0.0, 0.5, 0.0}

\newcommand{\set}[1]{\left\lbrace#1\right\rbrace}
\newcommand{\setIn}[1]{\mathbbm{1}_{\set{#1}}}
\newcommand{\Ind}[1]{\mathbbm{1}_{#1}}
\newcommand{\abs}[1]{\left\lvert#1\right\rvert}
\newcommand{\norm}[1]{\left\lVert#1\right\rVert}

\newcommand{\E}{\mathbb{E}}
\newcommand{\N}{\mathbb{N}}
\newcommand{\R}{\mathbb{R}}
\newcommand{\Z}{\mathbb{Z}}

\newcommand{\bx}{\mathbf{x}}

\newcommand{\cA}{\mathcal{A}}
\newcommand{\cB}{\mathcal{B}}
\newcommand{\cD}{\mathcal{D}}

\newcommand{\cF}{\mathcal{F}}
\newcommand{\cG}{\mathcal{G}}
\newcommand{\cH}{\mathcal{H}}

\newcommand{\cM}{\mathcal{M}}

\newcommand{\cV}{\mathcal{V}}

\newcommand{\cZ}{\mathcal{Z}}

\renewcommand{\ge}{\geqslant}
\renewcommand{\le}{\leqslant}

\newcommand{\iid}{\emph{i.i.d.}\ }

\newcommand{\Prb}{\mathrm{Pr}}

\newcommand{\DeCa}{\textsc{DecAFork}\xspace}

\newcommand{\AC}{\textsc{Average Crossing} }
\newcommand{\ac}{\text{AC}\xspace}
\newcommand{\ER}{\text{Erd\H{o}s--R\'enyi}\xspace}

\title{Random Walk Learning and the Pac-Man Attack}

\author{Xingran Chen, Parimal Parag, Rohit Bhagat, Zonghong Liu, and Salim El Rouayheb
\IEEEcompsocitemizethanks 	 
{
\IEEEcompsocthanksitem  Xingran Chen, Rohit Bhagat, Zonghong Liu, and Salim El Rouayheb are with Department of Electrical and Computer Engineering,  Rutgers University, Piscataway Township, NJ 08854, USA	(E-mail: 
xingranc@ieee.org, \{rb1395, zl304, sye8\}@scarletmail.rutgers.edu).
\IEEEcompsocthanksitem Parimal Parag is with the Department of Electrical Communication
Engineering, Indian Institute of Science, Bangalore, Karnataka 560012, India (E-mail: parimal@iisc.ac.in).

\IEEEcompsocthanksitem{This paper has been accepted by IEEE ISIT 2026.}
}
}

\begin{document}

\maketitle

\begin{abstract}
Random walk (RW)-based algorithms have long been popular in distributed systems due to low overheads and scalability, with recent growing applications in decentralized learning. However, their reliance on local interactions makes them inherently vulnerable to malicious behavior. In this work, we investigate an adversarial threat that we term the ``Pac-Man'' attack, in which a malicious node probabilistically terminates any RW that visits it. This stealthy behavior gradually eliminates active RWs from the network, effectively halting the decentralized operators without triggering failure alarms. To counter this threat, we propose the \AC (\ac) algorithm---a fully decentralized mechanism for duplicating RWs to prevent RW extinction in the presence of Pac-Man. Our theoretical analysis establishes that (i) the RW population remains almost surely bounded under \ac and (ii) RW-based stochastic gradient descent remains convergent under \ac, even in the presence of Pac-Man, with a quantifiable deviation from the true optimum. Furthermore, they uncover a phase transition in the extinction probability as a function of the duplication threshold, which we explain through theoretical analysis of a simplified variant of \ac. Our extensive empirical results on both synthetic and public benchmark datasets validate our theoretical findings.
\end{abstract}

\section{Introduction}\label{sec:Introduction}

Decentralized algorithms are becoming increasingly important in modern large-scale networked applications. 
These algorithms enable a network of nodes/agents, each with access only to local data and a limited local view of the connection graph, to collaborate in solving global computational tasks without centralized coordination. 
Among the most widely studied approaches are random-walk (RW)-based algorithms \cite{lovasz1993random, levin2017markov} and gossip-based algorithms \cite{pmlr-v119-koloskova20a}. 

While gossip-based methods are powerful, their repeated local broadcasts can lead to large communication overhead in large-scale systems \cite{pmlr-v119-koloskova20a}. 
This motivates the study of RW-based algorithms, which offer a communication-efficient alternative for large networks due to their simplicity and scalability \cite{lovasz1993random, levin2017markov}. As a result, RW-based methods have been successfully applied across a wide range of domains \cite{page1999pagerank, fouss2007random, liu2016smartwalk, backstrom2011supervised, zhang2019heterogeneous}.
A particularly compelling application is decentralized machine learning, where RWs are used to aggregate learning on data distributed across networked agents \cite{johansson2007simple, ayache2023walk}.

Despite their advantages, RW learning algorithms are  susceptible to security threats \cite{10693599, 9714881}.  We focus in this work   on a particular threat in which certain nodes behave maliciously by terminating, or ``killing'', any random walk that reaches them. We refer to this as the \textit{Pac-Man attack}, evoking the arcade character known for devouring everything in its path.


Recent work by~\cite{selfdup} introduced a novel approach based on \textit{self-regulating} RWs, aimed at enhancing the resilience of RW-based algorithms in a decentralized manner. The main algorithm proposed in this work is called \DeCa, which operates by maintaining a history of RW visit times at each node. The work in \cite{selfdup} provides theoretical guarantees for catastrophic failure scenarios under idealized assumptions and approximations. The followup work in \cite{egger2024self} includes numerical simulations evaluating the \DeCa’s behavior in the presence of malicious nodes.

In this work, to counter the Pac-Man attack, we propose a new decentralized duplication mechanism, termed the \AC (\ac) algorithm (see Algorithm~\ref{alg:AC}).  In the \ac algorithm, each benign node independently decides whether to duplicate an incoming RW based solely on local timing information—specifically, the time interval between two successive visits from any RW. If this interval exceeds a predetermined threshold, the node infers that at least one RW may have been killed by the Pac-Man and, with a given probability, duplicates the currently visiting RW. In the following, we summarize the main contributions of this article.

\noindent\textit{(i) Novel theoretical framework and analytical results}:
We develop a novel and rigorous theoretical framework to analyze the evolution of the number of active RWs under \ac, explicitly accounting for the strong interdependence among RWs. Within this framework, we establish that the RW population under \ac is almost surely bounded (Theorem~\ref{thm:FiniteRWs}).

Beyond boundedness, we further reveal a phase transition behavior of the RW population with respect to the duplication threshold. Specifically, for appropriately chosen thresholds, the RW population persists with positive probability, whereas overly large thresholds lead to almost sure extinction. To theoretically characterize this phenomenon, we introduce a simplified and analytically tractable variant of the AC algorithm, under which we rigorously establish the existence of a phase transition (see Proposition~\ref{pro:PhaseTrans}).

\noindent\textit{(ii) Integration with decentralized learning and proof of convergence}:
The \ac mechanism can be seamlessly integrated into random walk stochastic gradient descent (RW-SGD) \cite{johansson2007simple, sun2018markov, mao2020walkman}. 
We prove that RW-SGD converges under the \ac mechanism in the presence of an adversary (see Theorem~\ref{thm:Effectiveness}).  
We further characterize the bias induced by premature RW termination, showing how adversarial behavior skews the optimizer and leads to a bounded deviation from the true global optimum (Proposition~\ref{pro:Bounds}).

\section{Problem Formulation}\label{sec:ProblemFormulation}
\subsection{Graph and Random Walks}\label{subsec:GraphRW}

These agents are modeled as nodes in a {\it connected} undirected graph, where each agent possesses its own local data and can communicate only with neighboring agents.\footnote{Without loss of generality, we restrict our analysis to connected graphs. For disconnected graphs, the proposed theoretical framework can be applied separately to each connected component, and the analysis proceeds analogously.} Without loss of generality, we assume that the Pac-Man agent is indexed as node~$1$. 
The communication topology graph is constructed as a finite graph $\cG \triangleq (\cV, E)$, where the set of agents/nodes is $\cV\triangleq [N]$, and $E \subseteq \binom{\cV}{2}$ denotes the set of edges. In addition, let $\cB\triangleq\cV\backslash\set{1}$ denote the set of benign nodes.

In this system, computation is carried out via RWs on the graph. Each RW is determined by a fixed and identical transition probability matrix $P$. Moreover, each RW carries a token message (which can be viewed as a global task) that is processed and passed along the network. 
At each time step, only the node currently holding a token performs local computation and updates the message. After computation, the current node forwards the token to a randomly selected neighbors based on the transition probability matrix $P$. This process continues until a predefined stopping criterion is satisfied. 
When multiple RWs are present, each is uniquely identifiable (e.g., via an index $j$) to track their individual progress through the network.

\subsection{Threat Model: The Pac-Man Attack}\label{subsec:Pack-ManAttack}
We are interested in ensuring the resiliency of RW-based algorithms---that is, their ability to prevent total  extinction and continue operating effectively even when a malicious node attempts to terminate the RWs. Specifically, we focus on a threat we refer to as the \emph{Pac-Man attack}: a malicious node, termed \emph{Pac-Man}, terminates  all incoming RWs without performing the required computation or forwarding the result to a neighbor\footnote{In this paper, we focus on the case of a single Pac-Man node to simplify the presentation of the algorithm and its theoretical analysis. The proposed framework naturally extends to multiple Pac-Man nodes; see Remark~\ref{remark:MultiplePacMan} in Appendix~\ref{Appe:assumptions} for further discussion.}. This setting already captures the core difficulty of the problem: even one such adversary is sufficient to cause eventual extinction of all RWs with probability one, thereby completely halting the global task. 
The Pac-Man is particularly dangerous due to its ability to remain hidden through a deceptive behavior. 
\begin{compactenum}[(i)]
\item  The Pac-Man node can reply positively to all network-level fault-tolerance checks and controls, for example retransmissions and timeouts \cite{schneider1990implementing, lamport1998part, lamport1982byzantine, elnozahy2002survey, chandra1996unreliable}, making it difficult to distinguish it from benign nodes despite its malicious actions.
\item To avoid detection, the Pac-Man node terminates incoming RWs independently with a probability $\zeta \in (0, 1]$, referred to as the termination probability. This randomized behavior allows the Pac-Man to conceal itself among benign nodes. If the Pac-Man node terminates all incoming RWs, it would never propagate RWs to its neighbors, making it easily identifiable as malicious since no RWs would be observed from that node over a long time horizon. In contrast, by terminating incoming RWs only with a positive probability, the Pac-Man can intermittently forward RWs, thereby blending in with normal nodes. 
\end{compactenum}

One commonly used fault-tolerance technique is the introduction of redundancy.  However, static redundancy alone is {\it ineffective} in this adversarial setup. Simply starting with multiple RWs does not guarantee their survival: for any small termination probability $\zeta>0$, all RWs will eventually be terminated with probability $1$, causing the task to fail. Thus, redundancy cannot serve as a long-term solution.

\subsection{RW-based Stochastic Gradient Descent}\label{subsec:RWSGD}
We consider the following standard distributed optimization and learning problem:
\begin{align}\label{eq:goal0}
\min_{{\bf x}\in\R^m}f({\bf x}) = \min_{{\bf x}\in\R^m} \,\E_{u\sim\pi}\left[f_u({\bf x})\right],\,\,
\end{align}
where each node $u$  possesses a local function $f_u({\bf x})$, $\pi$ is a target sampling distribution, and $m\in\Z_+$. Here, we assume $[\pi]_u>0$ for $u\in\cV$.

Under local communication constraints, solving problem \eqref{eq:goal0} using RW-SGD algorithms has proven to be highly effective \cite{10.5555/3618408.3618786, doi:10.1137/08073038X, 10.5555/3327546.3327656}. 
Given a target sampling distribution $\pi$ and a connected graph $\cG$, The Metropolis–Hastings algorithm \cite{doi:10.1137/08073038X} constructs a RW on $\mathcal{G}$ with transition matrix $P$, for which $\pi$ is the stationary distribution.\footnote{The transition matrix $P$ depends on both the target sampling distribution $\pi$ and the graph $\cG$. We re-parameterize it as $P_{\pi,\cG}\in\R^{N\times N}$, and simply write $P$ when there is no risk of confusion.} The RW-SGD is outlined as ${\bf x}_{t+1} ={\bf x}_t - \gamma_t\hat{g}_{v_t}({\bf x}_t)$,
where $v_t$ is the node visited by the random walk at time $t$, and $\hat{g}_{v_t}({\bf x}_t)$ denotes the gradient or sub-gradient at that node. Here, the sequence $\{v_t\}_{t\in\N}$ evolves according to $P$.

\subsection{Designable Properties of the Decentralized Mechanism}\label{subsec:Objectives}
The entire goal is to design a \textit{decentralized mechanism} such that (i) the global task can be carried out via RW-based algorithms even in the presence of a Pac-Man node, and
(ii) when the proposed decentralized mechanism is integrated with RW-SGD, it remains effective and convergent.\footnote{In the absence of adversarial behavior, RW-SGD is known to converge to the global optimum \cite{10.5555/3618408.3618786, doi:10.1137/08073038X, 10.5555/3327546.3327656}. However, the introduction of a Pac-Man node fundamentally alters the dynamics. Since this adversarial node probabilistically terminates incoming RWs, it is no longer clear whether RW-SGD remains convergent.}
 
At the beginning of time slot $t$, let $\cZ_t$ denote the set of indices of active RWs, and define the random variable $Z_t \triangleq \abs{\cZ_t}$ as the total number of active RWs at that moment. For each RW $j \in \cZ_t$ at time $t$, we denote its location at time $t$ by $X_{j}(t) \in \cV$. At the initial time $t = 0$, we denote the number of initial RWs as $Z_0 = z_0$, where $z_0$ is a predetermined scalar. To achieve our entire goal, the proposed algorithm must have the following desirable properties.

\paragraph{\bf No Blowup} To ensure system stability, we must avoid uncontrolled growth in the number of RWs. The algorithm should keep the RW population bounded with probability one:
\begin{align}\label{eq:NoBlowup}
\Prb\left(\sup_t Z_t<\infty\,\middle|\, Z_0=z_0\right) = 1.
\end{align}

\paragraph{\bf Low Probability of Extinction}
To ensure the resiliency of the system, it is essential to sustain the RW population over time. The algorithm should keep that the probability that the Pac-Man attack eliminates all RWs is small. Let $\delta$ be a small threshold, we have
\begin{align}\label{eq:ExtinctionProb}
\Prb\left(\exists\, t_0,\,\forall\, t\ge t_0,\, Z_t=0 \,\middle|\, Z_0=z_0\right) < \delta.
\end{align}

\paragraph{\bf Convergence} 
All active RWs under RW-SGD must converge to the same minimizer, in the following sense: 
\begin{align}\label{eq:Effectiveness}
\lim_{t \to \infty} \E\left[\|\mathbf{x}^{(j_t)}_t - \tilde{\mathbf{x}}^\star\|\right]=0,\; \forall j_t\in\cZ_t,
\end{align}
where $\tilde{\mathbf{x}}^\star$ denotes the convergent point attained by the RW-SGD in the absence of a Pac-Man node. If the limit in \eqref{eq:Effectiveness} exists, we further aim to characterize the approximation error with respect to the true optimizer $\mathbf{x}^\star$, by bounding the deviation $\|\tilde{\mathbf{x}}^\star - \mathbf{x}^\star\|$.

\section{Average Crossing Algorithm}\label{sec:AC}

We present the \AC (\ac) algorithm, a decentralized mechanism for adaptively duplicating RWs to meet the designable properties outlined in Section~\ref{subsec:Objectives}.  We outline the  AC  algorithm in Algorithm~\ref{alg:AC}.

The core idea is as follows: each benign node $u \in \cB$ maintains a variable $L^{(u)}_t$, representing the most recent time \emph{prior to time $t$} at which node $u$ was visited by any RW.  If node $u$ has not been visited for too long, i.e., the interval $t-L_t^{(u)}$ exceeds a predetermined threshold $A_u$, node~$u$ suspects that at least one RW may have been lost. Then, with probability $q$, it duplicates a new RW as an {\it identical} copy of the currently visiting one (see lines~$5$-$6$ in Algorithm~\ref{alg:AC}). If multiple RWs arrive at node~$u$ at time~$t$ and the same condition met, node~$u$ selects one of the arriving RWs uniformly at random and duplicates it with probability $q$.

\begin{algorithm}
\caption{Average Crossing (AC) Algorithm}\label{alg:AC}
\begin{algorithmic}[1]
\State {\bf Input}: The graph $\mathcal{G}$, the thresholds $\{A_u\}_{u\in[N]}$, the initial recording $L_0^{(u)}=0$ for $u\in[N]$, the forking probability $q$, and the initial location of RWs $\cZ_0$.
\For{$t\ge 0$}
\For{$u\in\cB$}
\If{$u\in\cup_{j\in\cZ_t}X_{t, j}$}. 
\If{$u\in[N]$ and $t - L_t^{(u)} > A_u$}
\State With probability $q$, node~$u$ forks a new RW by uniformly copying one of the currently visiting RWs.
\EndIf
\State $L_t^{(u)}\leftarrow t$.
\EndIf
\EndFor
\EndFor
\end{algorithmic}
\end{algorithm}

\section{Fundamental Analysis}\label{sec:FundamentalAnalysis}

In this section, we present a rigorous theoretical analysis of the \ac algorithm, focusing on the designable properties outlined in Section~\ref{subsec:Objectives}. 
For brevity, the preliminaries---including notations, definitions, and assumptions---are deferred to Appendix~\ref{Appe:assumptions}.

Under the model assumption, the Pac-Man node is fixed at location $1$. We next present a formal definition of the system including the Pac-Man node. 
\begin{definition}\label{defn:PacMan}
Consider a communication topology $\cG = (\cV, E)$, as defined in Definition~\ref{defn:GenGph} (see Appendix~\ref{Appe:assumptions}), where node $1$ acts as the Pac-Man, such that if a RW visits node $1$, it is sent to a death node $0$\footnote{The death node is a virtual node used only for a clearer presentation.} with a termination probability $\zeta\in(0, 1]$. 
That is, we augment graph $\cG$ to $\cG^\prime = (\cV^\prime, E^\prime)$ where $\cV^\prime \triangleq \cV\cup\set{0}$ and $E^\prime \triangleq E \cup \set{(1,0)}$. 
In the presence of this Pac-Man, the original RW transition probability matrix $P$ now changes to $P^\prime$, where for each state $u,v \in \cV^\prime$
\begin{equation}
P^\prime_{uv} \triangleq 
\begin{cases}
P_{uv}, & u \in\cB, v \in \cV,\\
(1-\zeta)P_{uv}, & u =1, v \in \cV,\\
\zeta,& u = 1, v=0,\\
1,& u = 0, v = 0\\
0,&u=0, v\in\cV.
\end{cases}
\end{equation}
\end{definition}

\subsection{Population Boundedness}

First, we establish that the number of active RWs remains almost surely bounded at all times, despite ongoing duplications. This property is crucial for ensuring that the algorithm is stable and does not flood the network with RWs.

\begin{theorem}[Boundedness]\label{thm:FiniteRWs}
On \textbf{any} finite graph $\cG^\prime = (\cV^\prime, E^\prime)$ of Definition \ref{defn:PacMan}, 
with $z_0\ge 1$, $A_u \ge 1$, $q\le 1$, and $\zeta \in (0,1]$, the \ac algorithm ensures that 
$\limsup_{t\to\infty}Z_t < \infty$
almost surely.
\end{theorem}
\begin{proof}
The full proof is given in Appendix~\ref{Appe:FinitNum}. 
\end{proof}
Theorem~\ref{thm:FiniteRWs} implies that regardless of the thresholds $\{A_u\}_{u\in\cB}$ selected by the benign nodes, the RW population remains bounded over time almost surely. 

\subsection{Phase Transition in Extinction Behavior}
Second, we study how the duplication threshold affects whether the RW population survives or dies out. Intuitively, a small threshold leads to frequent duplications, increasing the chance of survival, while a large threshold delays responses and may result in extinction. Numerical experiments (Fig.~\ref{fig: behavior} in Section~\ref{sec: simulations} and Fig~\ref{fig: extinction more} in Appendix~\ref{Appe: simulations}) support this intuition, showing a soft phase transition in extinction probability as the threshold decreases. In particular, by choosing the duplication threshold appropriately, the extinction probability can be reduced to a very small positive value.

However, analyzing the \ac algorithm is challenging due to the implicit correlations in the duplication decisions across nodes. Upon each visit, node $u$ decides whether to duplicate by evaluating $t-L_t^{(u)}$, which reflects the cumulative influence of active RWs' trajectories from the initial time to the current moment. As a result, a benign node's duplication behavior is indirectly affected by the entire set of active RWs. These hidden correlations significantly complicate the analysis. 

To circumvent this difficulty and better understand the extinction behavior, we consider a simplified variant called the \textit{Weak Version of AC} (W-AC) algorithm, formally defined in Algorithm~\ref{alg:WAC} in Appendix~\ref{Appe:PhaseTrans}. The key difference is that, in W-AC, each RW decides autonomously when to duplicate based only on the time since its last visit to the current node, whereas in AC the duplication decision is made by the node.

\begin{proposition}[Soft Phase Transition]\label{pro:PhaseTrans}
Consider $z_0$ initial uniform RWs on an almost fully connected graph from Definition \ref{defn:AFCG} (in Appendix~\ref{Appe:PhaseTrans}).  
Then, under the W-AC algorithm defined in Algorithm~\ref{alg:WAC}, there exist two non-negative constants $\bar{\alpha}\ge \underline{\alpha}$, such that, for all $z_0 \ge 1$,
\begin{align*}
\left\{\begin{aligned}
&\Prb\left(\exists\, t_0,\,\forall\, t\ge t_0,\, Z_t=0 \,\middle|\, Z_0=z_0\right)=1&&\text{if }A\ge \bar{\alpha}\\
&\Prb\left(\exists\, t_0,\,\forall\, t\ge t_0,\, Z_t=0 \,\middle|\, Z_0=z_0\right)<1&&\text{if }A\le\underline{\alpha}.
\end{aligned}\right.
\end{align*}
\end{proposition}
\begin{proof}
The proof is given in Appendix~\ref{Appe:PhaseTrans}.
\end{proof}

\subsection{Convergence}
Now, we study how the \ac algorithm affects the convergence of RW-SGD in the presence of a Pac-Man node. 
It is important to note that each RW will eventually be terminated by the Pac-Man node with probability one. To address this, rather than relying on a single RW, we need to analyze the behavior of a {\it chain} of RWs. Specifically, whenever a RW duplicates, we refer to the original as the \textit{parent} and the resulting RW as its \textit{child}.

\begin{definition}[Chain of RWs]\label{defn:ChainRWs} 
Start with an initial RW $j_0\in[z_0]$. At each step $s\ge 0$, if $j_s$ has at least one child, let $j_{s+1}$ be a uniformly chosen child of $j_s$; otherwise, set $j_{s'}=-1$ for all $s'\ge s$ to mark termination. We refer to $\set{j_s}_{s\in\N}$ as a chain of RWs.
We call the chain \textit{infinite} if $\lim_{s\to\infty}j_s > 0$; otherwise, we call it \textit{finite}.
\end{definition}

\begin{figure}[htbp]
\centering
\includegraphics[width=0.9\linewidth]{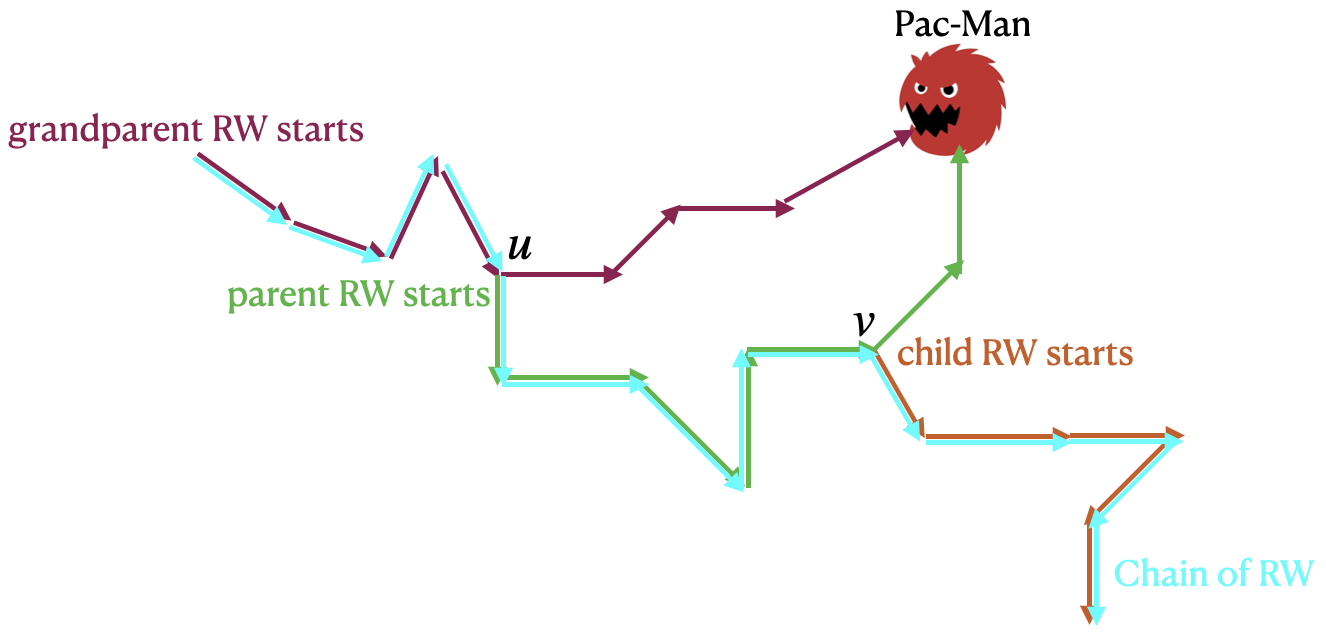}
\caption{Illustration of a chain of RWs. Node $u$ duplicates a new RW (green) by copying the purple RW. Node $v$ duplicates a new RW (orange) by copying the green RW. The blue trajectory shows how these RWs are connected over time, forming a chain.}
\label{fig:chain}
\end{figure}

We call a trajectory $\set{\cZ_t}_{t\in\N}$  \textit{surviving} if $\inf_{t\in\N}Z_t>0$. Under a surviving trajectory $\{\cZ_t\}_{t\in\N}$, there must exist at least one infinite chain of RWs; otherwise, 
if all chains of RWs are finite, then there exists some $t_0\in\Z$ such that $|Z_t|=0$ for $t\ge t_0$, which contradicts the definition of surviving trajectories.

\begin{theorem}\label{thm:Effectiveness}
Under the \ac mechanism and along a surviving trajectory $\set{\cZ_t}_{t \ge 0}$, RW-SGD converges to the minimizer of the following problem:
\begin{align}\label{eq:SurOpt}
\min_{{\bf x}\in\R^m}\tilde{f}({\bf x}) = \min_{{\bf x}\in\R^m} \,\E_{u\sim\pi^{(\zeta)}_{\text{chain}}}[f_u({\bf x})],
\end{align}	
where
\begin{align}\label{eq:pichain}
\pi^{(\zeta)}_{\text{chain}} = \left\{
\begin{aligned}
&[0, \nu^{(1)}]&&\zeta=1,\\
&\nu^{(\zeta)}&&\zeta\in(0,1),
\end{aligned}
\right.
\end{align}
and $\nu^{(\zeta)}$ is defined in Definition~\ref{def:matrixQ} in Appendix~\ref{Appe:assumptions}.
\end{theorem}
\begin{proof}
The full proof is given in Appendix~\ref{Appe:Effectiveness}. 
\end{proof}

We treat a chain of RWs as a {\it single effective} RW. The corresponding effective transition probability matrix $P^{(\zeta)}_{\text{chain}}$ is provided in  Appendix~\ref{Appe:UniqueP}. This matrix will be used to quantify the deviation of the new minimizer from the true global optimum $\bx^\star$. In the standard RW-SGD algorithm, when the stepsize $\eta_t$ decreases with time $t$ and tends to $0$, the algorithm converges to a deterministic point; if $\eta_t$ remains constant, it converges to a random variable \cite{10.5555/3618408.3618786, doi:10.1137/08073038X, 10.5555/3327546.3327656}.

\begin{proposition}[Shift of Optima]\label{pro:Bounds}
Let $P_{\mathrm{chain}}^{(\zeta)}$ be defined in Appendix~\ref{Appe:UniqueP}, with spectral gap $\gamma_\text{chain}^{(\zeta)}$. Let $\|\cdot\|_{\text{TV}}$ denote the total variation distance, and let ${\bx}_0$ be the starting point. Under the setting of Theorem~\ref{thm:Effectiveness}: 
\begin{compactenum}[(i)]
\item If the stepsize $\eta_t\downarrow 0$, then RW-SGD converges to the minimizer $\tilde{\bx}^\star$ of \eqref{eq:SurOpt}. Moreover, 
\begin{align*}
\frac{1}{L}\|\nabla f(\tilde{\bx}^\star\big)\|\le	\|\tilde{\bx}^\star-{\bx}^\star\|\le \frac{1}{\mu}\|\nabla f(\tilde{\bx}^\star)\|.
\end{align*}
\item If the stepsize $\eta_t=\eta < \tfrac{1}{L}$, then 
\begin{align*}
\E\left[\|\tilde{\bx}_T-{\bx}^\star\|\right]\le& 2(1-\eta\mu)^T\|{\bf x}_0 - {\bf x}^\star\|^2\\
+&\frac{\eta L\sigma^2}{\gamma_\text{chain}^{(\zeta)}\mu^2} + \frac{\left\|\tilde{\nu}^{(\zeta)} - \pi\right\|_{TV}^2\sigma^2L}{\mu^3}.
\end{align*}	
\end{compactenum}
\end{proposition}
\begin{proof}
The proof is given in Appendix~\ref{Appe:Bounds}.
\end{proof}

\section{Simulations}\label{sec: simulations}
In the main text, we only present representative simulations, and additional simulation results are provided in Appendix~\ref{Appe: simulations}.

\subsection{Simulation Setup}\label{subsec: Setup}
\paragraph{Graph Settings}
We consider a complete graph with $100$ nodes, consisting of one Pac-Man node and $99$ benign nodes. The target sampling distribution is uniform, i.e., $\pi = \left(\frac{1}{100}, \frac{1}{100}, \cdots, \frac{1}{100}\right)$. We assume $A_u = A$ for all $u \in \cB$. We set both the duplication and termination probabilities to $1$, i.e., $q = \zeta = 1$. We also consider three other connected graphs, including regular, ring, and \ER graphs; details are provided in Appendix~\ref{subsec:AdditionalGraphs}.

\paragraph{Learning Settings} For the distributed learning problem defined in \eqref{eq:goal0}, we evaluate our approach on both synthetic and public benchmark datasets. 

\textbf{Synthetic dataset}. We consider a decentralized linear regression task, where each node $u$ minimizes a local mean squared error (MSE) loss of the form:
$f_u({\bf x}) = ({\bf w}^T{\bf x}+b - y_u)^2$,
where ${\bf x}$ is the input feature, $y_u$ is the target label at node $u$, ${\bf w}$ is the weight vector, and $b$ is the bias. We assume each node holds only a single data point. This local objective $f_u({\bf x})$ is strongly convex and $L$-smooth.

\textbf{Public benchmark dataset}. We use the MNIST handwritten digit dataset \cite{deng2012mnist}. The dataset is evenly partitioned into $100$ disjoint subsets, with each node is assigned a unique subset. Each node $u$ minimizes the empirical cross-entropy loss over its local dataset $\cD_u$:
$f_u(w) = \frac{1}{|\cD_u|}\sum_{({\bf x}, y)\in\cD_u}\ell(w; {\bf x}, y)$,
where $w$ denotes the model parameters, $\ell(w; {\bf x}, y)$ is the cross-entropy loss function, and $\cD_u$ is the local subset assigned to node $u$. 
Both \iid and non-\iid data partitioning schemes for distributing data are considered; further details are provided in Appendix~\ref{subsec:partitioning}.\footnote{Results under \iid partitioning are provided in Appendix~\ref{subsec: convergence}, while the main text focuses on simulations under non-\iid partitioning.}

\paragraph{Baselines} To the best of our knowledge, the algorithm most closely related to our work is the \DeCa algorithm \cite{selfdup, egger2024self}. 
To ensure a fair comparison, \DeCa method adopts the same random walk transition matrix $P$, as used in the \ac algorithm, so their movement behavior remains identical.

Another baseline we consider is the classical \textsc{Gossip-based SGD} \cite{pmlr-v119-koloskova20a}. We incorporate the same adversarial setting as before. As shown in Fig.~\ref{fig:zetaregular} in Appendix~\ref{subsec:Gossip}, \textsc{Gossip-based SGD} exhibits significantly slow convergence in the Pac-Man setting. Consequently, we conclude that it is not a suitable baseline and exclude it from the remaining simulation results.

\subsection{Boundedness and Persistence}\label{subsec: Boundedness}

\begin{figure}[htbp]
  \centering
  \begin{minipage}[b]{0.48\linewidth}
    \centering
    \includegraphics[width=\linewidth, height=0.84\linewidth]{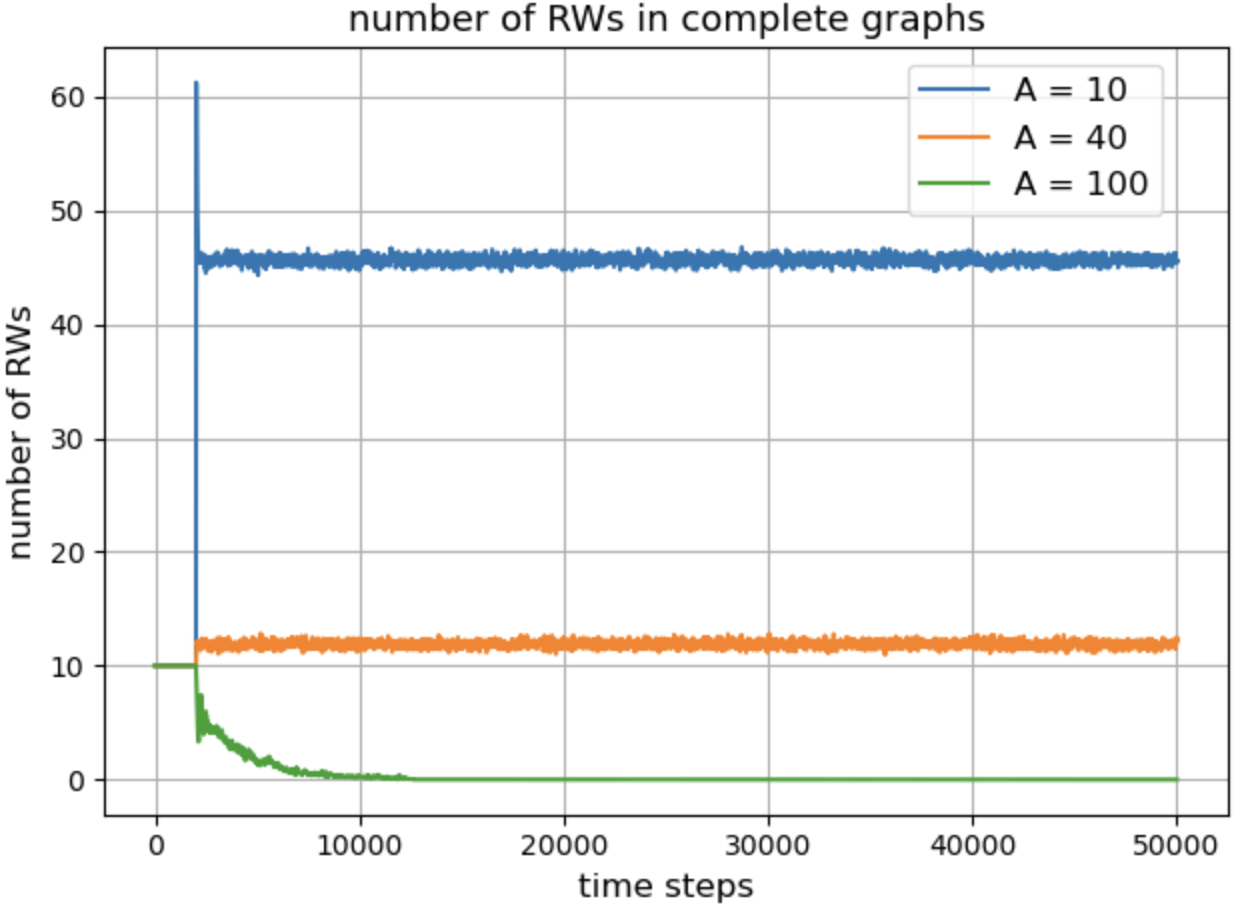}
    \caption*{(a) bounded population}
  \end{minipage}
  \hfill
    \begin{minipage}[b]{0.45\linewidth}
    \centering
    \includegraphics[width=\linewidth, height=0.9\linewidth]{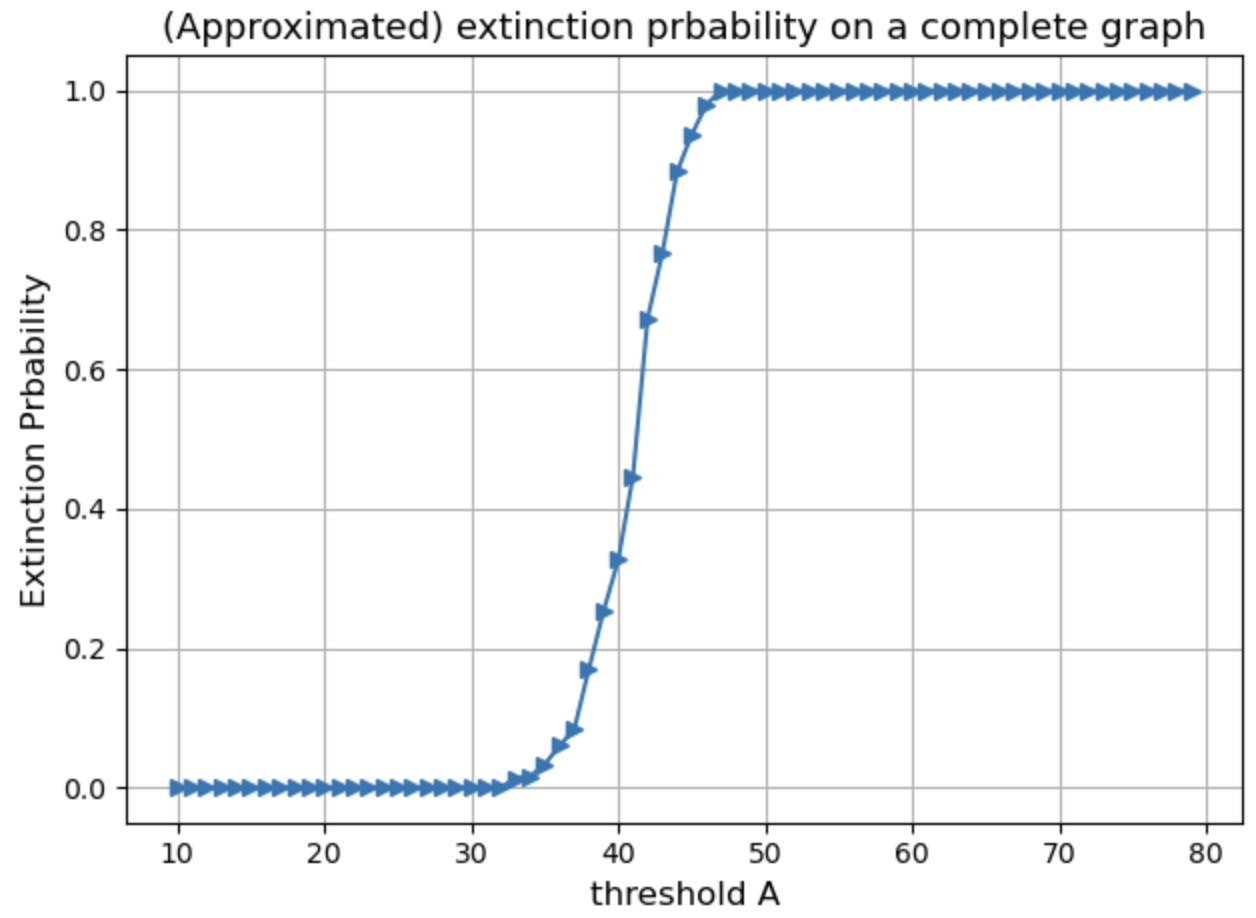}
    \caption*{(b) soft phase transition}
  \end{minipage}
  \caption{Behaviors of $\set{Z_t}_{t\ge0}$ on the complete graph.}
  \label{fig: behavior}
\end{figure}

We begin by validating Theorem~\ref{thm:FiniteRWs}. In Fig.~\ref{fig: behavior}(a), the $y$-axis shows the number of RWs, averaged over 100 iterations, while the $x$-axis denotes the time steps. We first allow the RWs to traverse without the presence of the Pac-Man for a short phase ($T_0 = 2000$). We then introduce the Pac-Man to run the \ac algorithm over a longer time horizon ($T = 50000$). The RW population process $\set{Z_t}_{t\in\Z}$ remains bounded over time. 

Then, we validate the persistence. Although Proposition~\ref{pro:PhaseTrans} focuses exclusively on the W-AC algorithm, we extend our investigation by simulating the \ac algorithm extinction behavior. Fig.~\ref{fig: behavior}(b) illustrates a soft phase transition in the extinction probability as a function of $A$. The $y$-axis shows the approximated extinction probability\footnote{We approximate the extinction probability by running a large number of simulations over a long time horizon and computing the ratio of runs in which the RW population goes extinct to the total number of runs.}, while the $x$-axis denotes the value of $A$. when $A$ exceeds a critical value (which depends on the graph topology), extinction occurs with probability $1$. Conversely, when $A$ falls below this critical threshold, the extinction probability drops sharply and approaches zero for small values of $A$. 

\subsection{Convergence}\label{subsec: convergence}

\begin{figure}[h]
\centering
  \centering
  \begin{minipage}[b]{0.45\linewidth}
    \centering
    \includegraphics[width=\linewidth, height=0.9\linewidth]{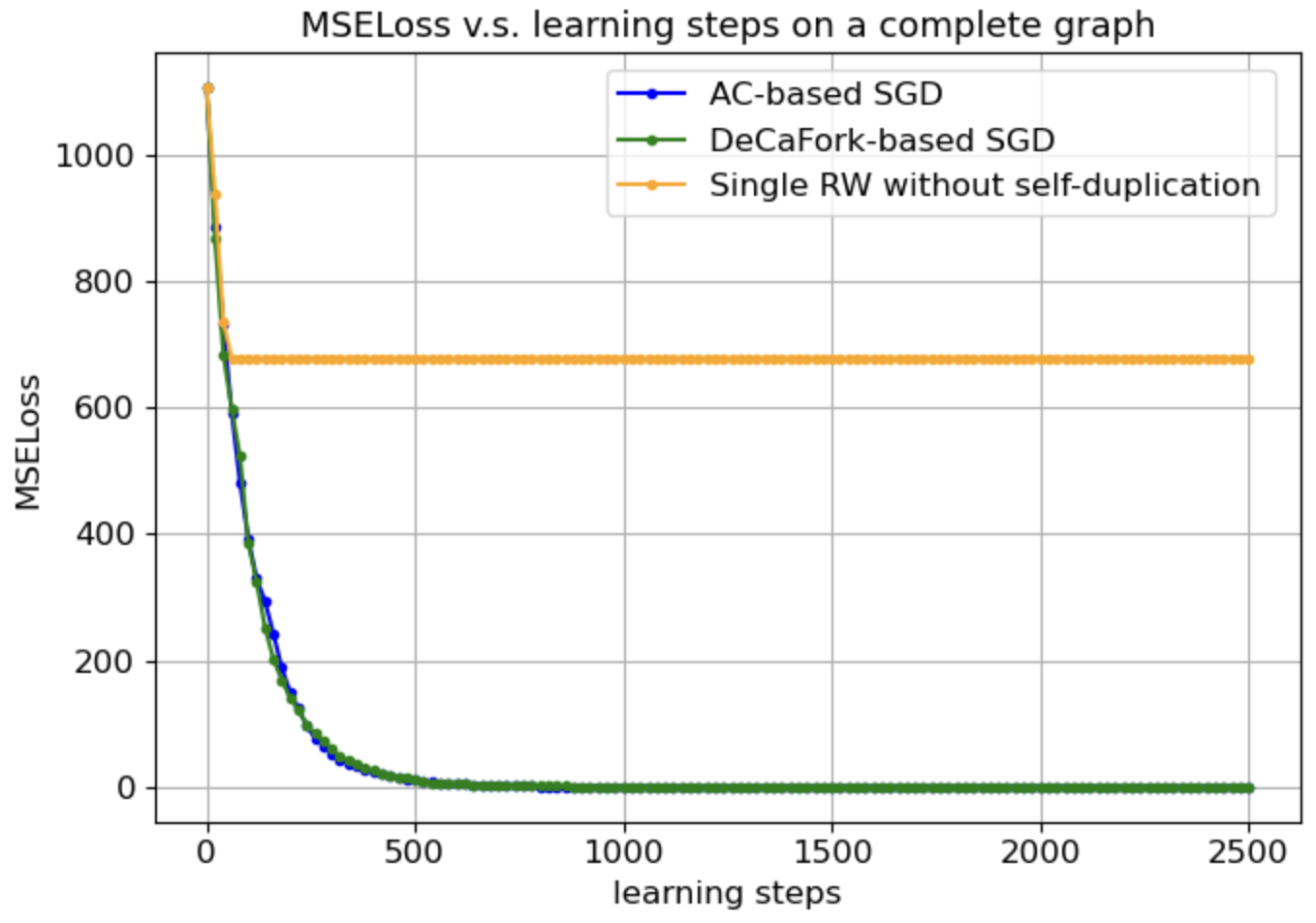}
    \caption*{synthetic dataset}
  \end{minipage}
  \hfill
  \begin{minipage}[b]{0.45\linewidth}
    \centering
    \includegraphics[width=\linewidth, height=0.9\linewidth]{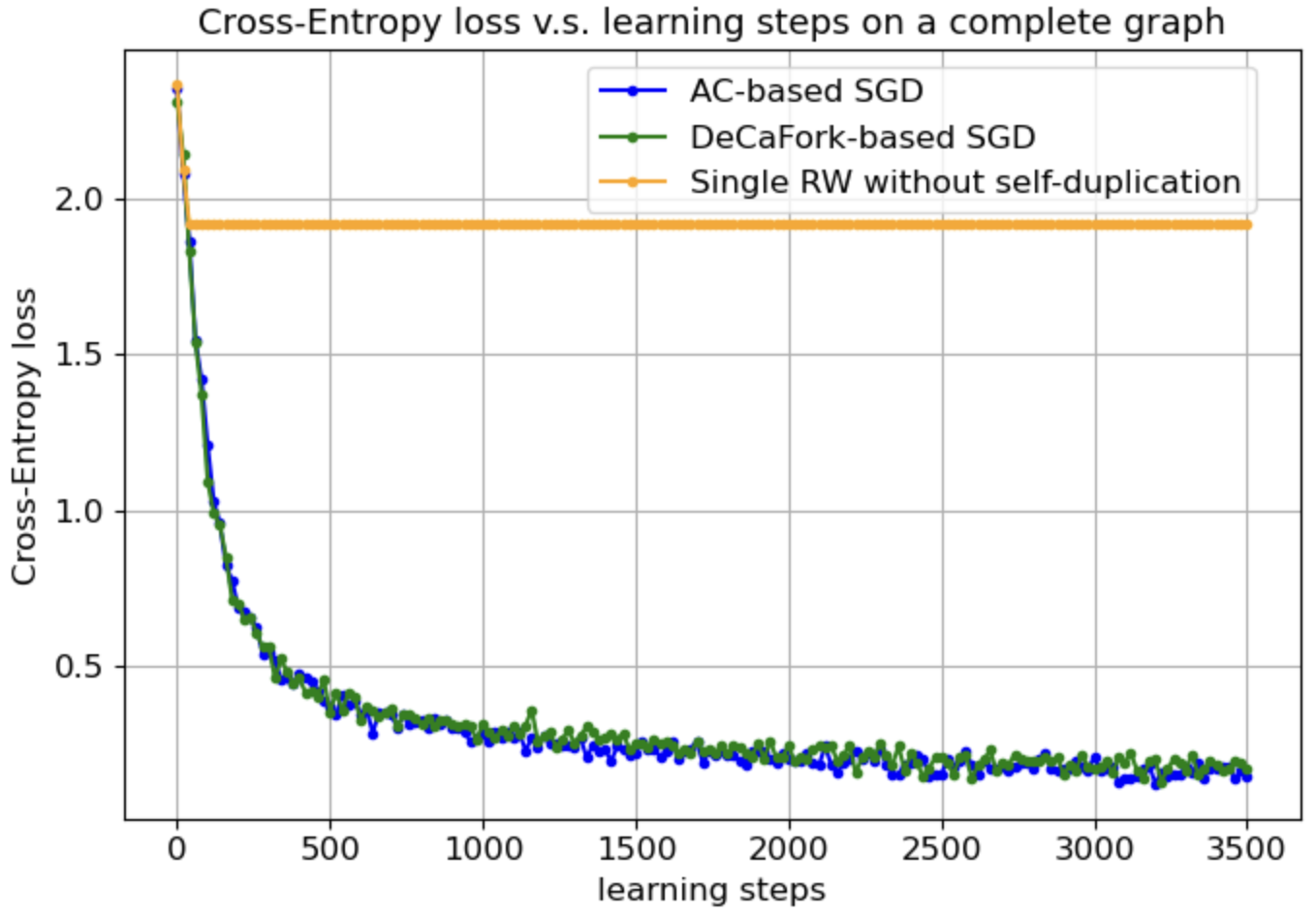}
    \caption*{public benchmark dataset}
  \end{minipage}
\caption{Loss function v.s. learning steps on different graphs.}
\label{fig:convergence}
\end{figure}

In Fig.~\ref{fig:convergence}, we present the convergence performance of RW-SGD under the \ac algorithm and compare it with two baselines: \DeCa and a single RW without self-duplication, using both synthetic and public benchmark datasets. The $y$-axis represents the loss curves, while the $x$-axis denotes the number of time steps. From the figure, both loss functions demonstrate effective convergence, which demonstrates Theorem~\ref{thm:Effectiveness}. Without self-duplication, the single RW terminates long before the learning process is completed, highlighting both the importance of self-duplication and the severity of the Pac-Man attack. Notably, the loss curves under RW-SGD are nearly identical across \ac and \DeCa. This is because, the two algorithms differ only in how RWs are duplicated, not in how active RWs move, their convergence behavior remains similar. 



\section{Conclusion}\label{sec:Conclusion}
In this work, we investigated the robustness of RW-based SGD under the Pac-Man attack. This stealthy threat gradually degrades the decentralized learning without triggering any  signals. To address this, we proposed the \ac algorithm, a fully decentralized duplication mechanism based solely on local visitation intervals, and provided rigorous analysis. Extensive experiments validate our findings.

\clearpage

\bibliographystyle{ieeetr}
\bibliography{references}

\clearpage
\appendices

\section{Preliminaries: Notations, Definitions, and Assumptions}\label{Appe:assumptions}
In this Appendix, we introduce the notation, definitions, and assumptions used in the system model and its transition dynamics.

\begin{definition}[Communication topologies and RWs]\label{defn:GenGph}
A communication topology is defined as a finite directed graph $\cG \triangleq (\cV, E)$ with the set of nodes $\cV = [N]$ and the set of edges $E \subseteq \binom{\cV}{2}$. 
Each RW $X_j:\Omega\to\cV^{\Z_+}$ on this graph is assumed to be \iid and can be defined by the common transition probability matrix $P:\cV\to \cM(\cV)$, 
where the probability of transition from node $u$ to node $v$ in one time step at time $t\in\Z_+$, is 
\begin{align*}
P_{uv} \triangleq \Prb\left(X_j(t+1)= v \,\middle|\, X_j(t)=u\right).
\end{align*}
We will call the RW on graph $\cG$ aperiodic if transition probability matrix $P$ is aperiodic. 
\end{definition}
\begin{remark}
Without loss of generality, we assume that $P_{uv} > 0$ for all nodes $v$ connected to node $u$ in graph $\cG$.  
Therefore, transition matrix $P$ is irreducible iff graph $\cG$ is connected. 
\end{remark}

\begin{definition}[Timing conventions]\label{defn:Timing} 
We define the timing conventions of the system as follows:
\begin{compactenum}[(1)]
\item At the beginning of time slot $t$, let $\cZ_t$ denote the set of indices of active RWs. 
Each active RW $j\in\cZ_t$ is associated with a birth time $\theta_j\ge0$ and an initial location $u_j \triangleq X_j(\theta_j)\in\cV$. 
Specifically, let $Z_t=\abs{\cZ_t}$.    
\item At the end of time slot $t$, each active RW $j\in\cZ_t$ moves from its current location to a randomly selected neighbor in the next time slot. 
We denote by $X_j(t+1)$ the location of RW $j$ after this movement at time $t+1$. 
\item Upon arrival at location $X_j(t+1)$ at time $t+1$, termination operation at node $1$ and creation operations at other nodes is performed (if applicable).  
\end{compactenum}
\end{definition}


Based on Definition~\ref{defn:PacMan}, the transition matrix $P'$ can be characterized by two cases:
$\zeta=1$ and $0<\zeta<1$. When $\zeta=1$, the Pac-Man node becomes an absorbing state, whereas when $0<\zeta<1$, the Pac-Man node is non-absorbing. In particular,
\begin{compactenum}[(1)]
\item When $\zeta=1$, the Pac-Man node eats every incoming RW, and the local data at the Pac-Man node cannot be utilized, making it an absorbing state. 
From Definition \ref{defn:PacMan}, the death node $0$ is also absorbing. 
Therefore, we merge the Pac-Man node and the death node into a single absorbing one, which we continue to denote as node $1$. 
The corresponding transition matrix $P^\prime$ becomes 
\begin{align}\label{eq:NewTransMatCase1}
P^\prime=\scalebox{0.8}{
$\begin{bmatrix}
1  & 0&\cdots& 0 \\
P_{21}&P_{22}&\cdots&P_{2N}\\
\vdots&\vdots&\vdots&\vdots\\
P_{N1}&P_{N2}&\cdots&P_{NN}
\end{bmatrix}$}
\triangleq\begin{bmatrix}
1& {\bf 0}_{1\times (N-1)}\\
R^{(1)}&Q^{(1)}
\end{bmatrix}.
\end{align}
We observe that $Q^{(1)} \in [0,1]^{\cB\times\cB}$ is sub-stochastic matrix where $Q^{(1)}_{uv} = P_{uv}$ for each $u,v \in \cB$, and $R^{(1)} \in [0,1]^{\cB\times 1}$ column vector where $R^{(1)}_u = P_{u1}$ for each benign node $u \in \cB$. 
\item When $0<\zeta<1$,  the local data at the Pac-Man node cannot be reliably utilized, because any RW visiting it is terminated with probability strictly less than $1$.
In this case, only the death node $0$ is absorbing. 
For clarity, we denote the location of the death node as $0$. 
According to Definition \ref{defn:PacMan} and denoting $\bar\zeta \triangleq 1-\zeta$, the corresponding transition matrix $P^\prime$ becomes
\begin{align}\label{eq:NewTransMatCase2}
P^\prime=&\scalebox{0.8}{
$\begin{bmatrix}
1  &0 &\cdots& 0 \\
\zeta &\bar\zeta P_{11}&\cdots&\bar\zeta P_{1N}\\
0&P_{21}&\cdots&P_{2N}\\
\vdots&\vdots&\vdots&\vdots\\
0&P_{N1}&\cdots&P_{NN}
\end{bmatrix}$}\triangleq\begin{bmatrix}
1& {\bf 0}_{1\times N}\\
R^{(\zeta)} & Q^{(\zeta)}
\end{bmatrix}.
\end{align}
We observe that 
\begin{xalignat*}{2}
&R^{(\zeta)} \triangleq \begin{bmatrix}\zeta\\  {\bf 0}_{(N-1) \times 1} \end{bmatrix},&&Q^{(\zeta)} \triangleq \begin{bmatrix}\bar\zeta P_{11}& \bar\zeta S\\ R^{(1)} & Q^{(1)}\end{bmatrix}.
\end{xalignat*} 
The row vector $S \in [0,1]^{1\times \cB}$ such that $S_{1v} = P_{1v}$ for each benign node $v \in \cB$. 
\end{compactenum}
\begin{remark}\label{remark:MultiplePacMan}
If $\zeta=1$, the analysis extends straightforwardly to the setting with multiple Pac-Man nodes. In this case, all Pac-Man nodes can be treated as a single ``super node'', which acts as an absorbing state of the system. The resulting analysis is essentially identical to that of the single Pac-Man node case. 
If $\zeta<1$, the analysis becomes more involved, since the transition probability to the absorbing state (i.e., the death node $0$) depends on the identity of the Pac-Man node through (a) their corresponding termination probabilities and (b) their connectivity to the benign nodes.
Nevertheless, the analysis can still be carried out within the theoretical framework proposed for the single Pac-Man node setting, although the analysis would be more tedious for a larger number of Pac-Man nodes. 
\end{remark}
\begin{definition}[Robustly Connected Graph]\label{defn:StrongConnect}
A graph $\mathcal{G}$ is robustly connected if
\begin{compactenum}[(i)]
\item every pair of benign nodes in $\cB$ is connected by a path that avoids the Pac-Man node, and
\item the Markov chain corresponding to each active RW is aperiodic.
\end{compactenum}
\end{definition}
\begin{remark}
According to Definition~\ref{defn:StrongConnect}, a robustly connected graph cannot be partitioned into two disjoint components by Pac-Man. In addition, in a robustly connected graph, the Markov chain corresponding to each active RW is irreducible and aperiodic.
\end{remark}

\begin{definition}\label{def:matrixQ}
For $\zeta \in (0,1]$, let $Q^{(\zeta)}$ be the matrix defined in \eqref{eq:NewTransMatCase1} and \eqref{eq:NewTransMatCase2}.  
Let $\alpha^{(\zeta)}$ denote the (unique) maximum eigenvalue of $Q^{(\zeta)}$\footnote{
We assume a unique maximum eigenvalue for simplicity.  
If $Q^{(\zeta)}$ has multiple dominant eigenvalues, the arguments can be extended using a standard Jordan decomposition.
},  
and let 
$\nu^{(\zeta)}$ denote the associated 
left normalized positive eigenvectors with unit sum.  
\end{definition}

\begin{assumption}\label{assu:IndependentRW}
Each RW has an \iid evolution on this graph with transition probability matrix $P$, conditioned on their initial locations. 
\end{assumption}

\begin{assumption}\label{assu:robustlyconnected}
The graph $\cG$ defined in Definition~\ref{defn:GenGph} is a robustly connected graph.
\end{assumption}
Next, we adopt the common assumptions used in standard distributed optimization problems, as follows.
\begin{assumption}\label{assu:StandCons}
Each local function $f_u({\bf x})$ in \eqref{eq:goal0} with $u\in\cV$ is $\mu$-strongly convex and $L$-smooth.
\end{assumption}

\begin{assumption}\label{assu:Boundedgrad}
Bounded norm of the local gradient at the global optimum ${\bx}^\star$, i.e. 
$
\sup_{u \in \cV}\norm{\nabla f_u({\bx}^\star)}^2\le \sigma^2,
$
where ${\bx}^\star$ is the minimizer of \eqref{eq:goal0}.
\end{assumption}
In the remainder of this article, we assume that Assumptions \ref{assu:IndependentRW} -- \ref{assu:Boundedgrad} hold.

\section{Proof of Theorem~\ref{thm:FiniteRWs}}\label{Appe:FinitNum}
\begin{definition}\label{defn:Initdistribution}
Let $\Prb_u$ denote the probability measure under which the RW $j$ starts at node $u$, i.e., $X_{0, j}=u$. Given a distribution $\nu$ over the node set $[N]$, we define the mixed law $\Prb_{\nu}$
\begin{align*}
\Prb_{\nu}=\int_{u\in [N]}\Prb_u d\nu(u),
\end{align*}
which corresponds to initializing RW $j$ according to $\nu$.
\end{definition}

\begin{definition}\label{defn:NaturalFilt}
Consider the number of active RWs $Z_t$ at time $t \in \Z_+$. The natural filtration for the random sequence $Z \triangleq (Z_t: t \in \Z_+)$ is denoted by $\cF_\bullet \triangleq (\cF_t: t \in \Z_+)$ where $\cF_t \triangleq \sigma(Z_s, s \le t)$. 
\end{definition}

Consider the graph $\cG^\prime$ defined in Definition \ref{defn:PacMan}. Since the original graph $\cG$ is connected and finite, then the Pac-Man is {\it reachable} from any other node $u \in [N]$, i.e., there exists $n_u \in \N$ such that $\left(P_{u0}^\prime\right)^{n_u} > 0$.
\begin{definition}
\label{defn:MinPathPacMan} 
We define the smallest number of steps to reach the Pac-Man from node $u\in[N]$ as 
\begin{align*}
d_u \triangleq \inf\set{n \in \N: \left(P_{u0}'\right)^n > 0}.    
\end{align*}
We define the maximum of the minimum time steps to reach the Pac-Man from node $u \in [N]$ as 
\begin{align}\label{eq:MaxMinPath}
d \triangleq \max_{u \in [N]}d_u.    
\end{align}
Accordingly, we define the smallest probability of reaching the Pac-Man within $d$ steps from node $u \in [N]$, as 
\begin{align*}
c \triangleq \min_{u \in [N]}\left(P_{u0}'\right)^{d_u}. 
\end{align*}
\end{definition}
Since the Pac-Man is reachable, $n_u$ is finite for each $u\in[N]$, then $d_u \le n_u$ and hence is finite. By Definition~\ref{defn:PacMan}, $P^\prime_{u0} >0$ for each $u \in [N]$, and then $c$ is positive.  Thus, $d$ and $c$ are well-defined.

\begin{definition}\label{defn:BirthDeathRW}
Consider a finite connected graph $\cG$ with $d$ defined in Definition~\ref{defn:MinPathPacMan}. 
During a fixed finite and half-open time interval $T \subseteq \R_+$, we denote the number of random walks that hit PacMan by $D_T$ and the number of RWs generated by $G_T$.  
\end{definition}

\begin{lemma}
Consider $Z_t$ independent, aperiodic, active RWs at time $t$, each following the identical law over a finite connected graph $\cG$ with $d$ defined in \eqref{eq:MaxMinPath}. 
Then, 
\begin{align}\label{eqn:MeanRWUB}
\E[Z_{t+d}-Z_t\mid\cF_t] \le  - c\zeta Z_t + (N-1)d.
\end{align} 
\end{lemma}
\begin{IEEEproof} 
From the definition of $D$ and $G$ from Definition \ref{defn:BirthDeathRW}, we can write the difference in the number of active RWs at time $t+d$ and $t$ as 
\begin{align}\label{eqn:DiffRW}
Z_{t+d} = Z_t - D_{(t, t+d]} + G_{(t, t+d]}.
\end{align}
We note that the graph $\cG$ has $(N-1)$ benign nodes, and at most one RW can be generated at each node at each time $t$, 
Thus, we have
\begin{align}\label{eqn:BirthRW}
\E[G_{(t, t+d]}\mid \cF_t] \le \sum_{s=0}^{d-1}\left(Z_{t+s}\wedge (N-1)\right) \le (N-1)d.
\end{align}
At time $t$, there are $Z_t$ active RWs. 
From Definition~\ref{defn:MinPathPacMan}, for any RW $X_j$ with transition probability matrix $P$, we have
\begin{align*}
\Prb_u\left(\cup_{n=1}^d\set{X_{n, j} = 1}\right) \ge \Prb_u\left(X_{d_u, j} = 1\right) \ge c,
\end{align*}
where $\Prb_u$ is defined in Definition~\ref{defn:Initdistribution}.
That is, $c$ is the uniform lower bound on the probability of ending up at the Pac-Man \emph{within} $d$ steps, over all possible initial positions. 
It follows that the number of deaths for RWs is lower bounded by the number of active RWs at time $t$ hitting Pac-Man (ignoring the RWs generated during this interval and hitting Pac-Man), and hence 
\begin{align}\label{eqn:DeathRW}
\E[D_{(t,t+d]}\mid \cF_t] \ge c\zeta Z_t.
\end{align}
Taking conditional expectation of \eqref{eqn:DiffRW} given history $\cF_t$, substituting the upper bound on the conditional mean number of births \eqref{eqn:BirthRW} and the lower bound on the conditional mean number of deaths \eqref{eqn:DeathRW}, we obtain the result. 
\end{IEEEproof}
\begin{cor}\label{cor:Foster-Lyapunov} 
Consider independent aperiodic RWs on a finite connected graph $\cG$ with identical probability laws and $d$ defined in \eqref{eq:MaxMinPath}. 
For any $\epsilon > 0$, there exists positive constants $b, B$, such that $B \ge b$ and the random sequence $Z$ satisfies the following conditions.
\begin{compactenum}[(a)]
\item If $Z_{t}\le B$, then $\E[Z_{t+d}\mid\cF_{t}]\le b$.
\item If $Z_{t} > B$, then $\E[Z_{t+d}-Z_{t}\mid\cF_{t}]<-\epsilon$. 
\end{compactenum}
\end{cor}
\begin{IEEEproof}
Let $\epsilon > 0$ and $N$ be the number of nodes in $\cG$.  
We define $B \triangleq \frac{1}{c\zeta}\left((N-1)d+\epsilon\right)$ and $b \triangleq (1-c\zeta)B+ (N-1)d$ for $c$ defined in Definition~\ref{defn:MinPathPacMan}. 
It follows that $b, B$ are positive and $b \le B$. 
\begin{compactenum}[(a)]
\item Let $Z_t \le B$. 
It follows from \eqref{eqn:MeanRWUB}, that
\begin{align*}
\E[Z_{t+d}\mid \cF_t]\le (1-c\zeta)B + (N-1) d = b. 
\end{align*}
\item Ley $Z_t > B$. It follows from \eqref{eqn:MeanRWUB} and definition of $B$, that 
\begin{align*}
&\E[Z_{t+d}-Z_t\mid\cF_t] \le - c\zeta Z_t + (N-1)d \le -\epsilon.
\end{align*}
\end{compactenum}
\end{IEEEproof}

\begin{definition}[Supermartingale] 
\label{defn:SupMart}
Consider independent aperiodic RWs on a finite connected graph $\cG$ with identical probability laws and $d$ defined in \eqref{eq:MaxMinPath}, $\epsilon > 0$, and positive constants $b, B$ defined in Corollary~\ref{cor:Foster-Lyapunov}. For $t_0\ge 0$ and $k \in \Z_+$, we define periodic samples of number of active RWs and its natural filtration at time $t_0+dk$ as 
We define a Lyapunov function $V:\R_+\to\R_+$ for each random variable $Z\in \R_+$
\begin{align}\label{eqn:Lyapunov}
V(Z) \triangleq Z\Ind{\set{\E[Z]>B}} + B\Ind{\set{\E[Z]\le B}}. 
\end{align}
For $t_0\ge 0$ and $k \in \Z_+$, we define periodic samples of number of active RWs and its natural filtration at time $t_0+dk$ as  
\begin{xalignat*}{2}
&M_k \triangleq V(Z_{t_0+dk}),&&\cH_k \triangleq \cF_{t_0+dk}.
\end{xalignat*}
We define a random sequence $M \triangleq (M_k: k \in \Z_+)$ and filtration $\cH_\bullet \triangleq (\cH_k:k\in \Z_+)$. 
\end{definition}
\begin{lemma}\label{lem:SupMart}
Sequence $M$ is a supermartingale adapted to filtration $\cH_\bullet$.  
\end{lemma}
\begin{IEEEproof}
We first observe that $Z_t$ is $\cF_t$ measurable, and hence $M_k$ is a $\cH_k$ measurable by definition. For each $k \in \Z_+$, we can define the following $\cH_k$ measurable events
\begin{align*}
&\cA_{k,1} \triangleq \set{Z_{t_0+dk}=z_{t_0+dk}\le B, \E[Z_{t_0+dk}] \le B},\\
&\cA_{k,2} \triangleq \set{Z_{t_0+dk} =z_{t_0+dk} > B, \E[Z_{t_0+dk}] \le B},\\
&\cA_{k,3} \triangleq \set{Z_{t_0+dk}=z_{t_0+dk} \le B, \E[Z_{t_0+dk}] > B},\\
&\cA_{k,4} \triangleq \set{Z_{t_0+dk} =z_{t_0+dk} > B, \E[Z_{t_0+dk}] > B}.
\end{align*}
In terms of these events, we can write the conditional mean as 
\begin{align*}
&\E[(M_{k+1}-M_k)\mid\cH_k] \\
&= \E[(M_{k+1}-M_k)(\Ind{\cA_{k,1}}+\Ind{\cA_{k,2}}+\Ind{\cA_{k,3}}+\Ind{\cA_{k,4}})\mid\cH_k]. 
\end{align*}
Let us compute $\E[(M_{k+1}-M_k)\Ind{\cA_{k,1}}\mid\cH_k]$ as an example: from Corollary~\ref{cor:Foster-Lyapunov}  and definition \eqref{eqn:Lyapunov},
if $\Ind{\cA_{k,1}}=1$, then $z_{t_0+dk}\le B$, thus $\E[Z_{t_0+d(k+1)}\mid\cH_k] = b < B$, we have $M_{k+1} = B$. In addition, since $\Ind{\cA_{k,1}}=1$, then $\E[Z_{t_0+dk}]\le B$, thus $M_{k} = B$, and 
\begin{align*}
\E[(M_{k+1}-M_k)\Ind{\cA_{k,1}}\mid\cH_k] = B - B = 0.
\end{align*}
Similarly, we have:
\begin{align*}
&\E[(M_{k+1}-M_k)\Ind{\cA_{k,2}}\mid\cH_k] < -\epsilon,\\
&\E[(M_{k+1}-M_k)\Ind{\cA_{k,3}}\mid\cH_k] < 0,\\
&\E[(M_{k+1}-M_k)\Ind{\cA_{k,4}}\mid\cH_k] < -\epsilon.
\end{align*}
Combining the these results, we get the result.
\end{IEEEproof}

\subsection{Proof of Theorem~\ref{thm:FiniteRWs}}

We define a stopping time $\tau_0 \triangleq \inf\set{t \in \Z_+: Z_t = 0}$. If $\tau_0<\infty$, then $Z_t = 0$ for all $t \ge \tau_0$, which implies $\limsup_{t\to\infty}Z_t<\infty$.

Therefore, without loss of generality, we consider the case when $Z_t > 0$ for any finite time $t$. 
From the definition of sequence $M$ and filtration $\cH_\bullet$ in Definition \ref{defn:SupMart}, Lemma \ref{lem:SupMart}, and positivity of $Z$, we observe that $M$ is a positive supermartingale adapted to filtration $\cH_\bullet$. 
By the Doob’s supermartingale convergence Theorem \cite{bookprobability}, supermartingale $M$ converges to a limit $M_\infty$ almost surely, i.e.  
\begin{align*}
\lim_{k\to\infty}M_k = M_\infty <\infty,\,\,a.s.
\end{align*}
From the definition of supermartingale $M$ in Definition \ref{defn:SupMart}, it follows that for any $t_0\ge 0$, 
\begin{align*}
\limsup_{k\to\infty}Z_{t_0+d k}\le \max\set{B, M_\infty}<\infty.
\end{align*}
Since the choice of $t_0 \in \Z_+$ was arbitrary, we have 
\begin{align*}
\limsup_{t\to\infty}Z_t <\infty,\,\,a.s.
\end{align*}

\section{Soft Phase Transition in Complete Graphs}\label{Appe:PhaseTrans}
\subsection{Definitions and the Weak Version of AC Algorithm}
\begin{definition}[Almost fully connected graph $\cG$] 
\label{defn:AFCG}
Let the killing and forking probabilities $\zeta=q=1$. 
Consider a finite graph $\cG^\prime$ from Definition \ref{defn:PacMan} with a single malicious node $1$, and edges 
\begin{align*}
E =\binom{[N]}{2}\cup\set{(1,d)}\setminus \set{(1,n): n \ge 1}.
\end{align*}
We call this an \emph{almost fully connected graph}, and a RW on this graph with uniform transition probability\footnote{i.e. $P_{uv} = \frac{1}{N}$ for all $u\in\cB$ and $v \in \cV$.} is called \emph{uniform RW} on the almost fully connected graph. 
We assume that the initial location of each initial uniform RWs on this graph is sampled \iid uniform from all nodes $\cV$, and the duplication threshold at each benign node is identically $A_u = A \ge 1$. 
\end{definition}

\begin{definition}[Lifetime]\label{defn:StoppingTj}
Consider a RW $j\in\cZ_t$ with the birth time $\theta_j$ and the initial location $u_j\in\cB$. We define its lifetime as the first time it visits the Pac-Man node, given by
\begin{align}\label{eq:StoppingTj}
K^{(j)}=\inf\set{t\ge\theta_j: X_{t, j}=0,X_{\theta_j, j}=u_j}.
\end{align}
\end{definition}
The distribution of $K^{(j)}$ is independent of the initial location $u_j$ because the graph structure is complete and the transition matrix is uniform.

\begin{definition}[Age of Visiting]\label{defn:AgeofVisiting}
Let $L_t^{(u, j)}$, for $u\in\cB$ and $t\in[\theta_j, K^{(j)}]$, denote the most recent time up to $t$ at which RW $j\in\cZ_t$ visited node $u$:
\begin{align}\label{eq:LstVst}
L_t^{(u, j)} = \max\set{t'\in[\theta_j, t]: X_{t', j}=u, X_{\theta_j, j}=u_j}.
\end{align}
We define the {\it age of visiting (AoV)} of node~$u$ with respect to RW $j\in\cZ_t$ at time $t$, denoted $H_t^{(u,j)}$, as follows:
\begin{compactenum}[(1)]
\item if RW $j$ visits node $u$ at time $t$ for the first time, then
\begin{align}\label{eq:AoVuj0}
H_t^{(u, j)}=0;
\end{align}
\item if node~$u$ was visited by RW $j$ at least once before time $t$, then 
\begin{align}\label{eq:AoVuj1}
H_t^{(u,j)}=t-L_{t-1}^{(u,j)}.
\end{align} 
\end{compactenum}
Both $L_t^{(u, j)}$ and $H_t^{(u,j)}$ are defined only after the active RW $j\in\cZ_t$ has visited node~$u$. If RW $j$ has not yet visited node $u$, or has already been terminated, then $L_t^{(u, j)}$ and $H_t^{(u,j)}$ are assigned a null value.
\end{definition}
We specify the AoV of a newly generated RW $j$ at node $u_j$ at time $t$, following Definition~\ref{defn:Timing} and Definition~\ref{defn:AgeofVisiting}. Suppose that RW $j$ is generated at node $u_j$ at time $t$, we have $j\in\cZ_{t+1}$, and it begins its movement in the next time slot, i.e., at time $t+1$. As the RW $j$ is initialized at node $u_j$ at time $t$,  we define the AoV at that moment as $0$, i.e., $H_t^{(u_j, j)} =0$.

Instead of analyzing the full AC algorithm, we focus on a simplified variant known as {\it weak version of AC (W-AC)} algorithm, which is outlined below.
\begin{algorithm}
\caption{Weak Version of AC (W-AC)}\label{alg:WAC}
\begin{algorithmic}[1]
\State {\bf Input}: The graph $\mathcal{G}$, the threshold $A\ge 0$, the forking probability $q=1$, and the indices of initial RWs $\cZ_0$.
\For{$t\ge 0$}
\For{$j\in\cZ_t$}
\State If $X_{j, t}=u\in\cB$ and $H_{t}^{(u, j)}\ge A$, RW $j$ forks a new RW, denoted by $j'$, as an identical copy of it. Set $H_t^{(u, j')}=0$ and $\cZ_{t+1}=\cZ_t\cup\set{j'}$; otherwise, if $H_{t}^{(u, j)} < A$, RW $j$ is moved to one of the neighbors of node~$u$ uniformly. 
\EndFor
\EndFor
\end{algorithmic}
\end{algorithm}

\subsection{Proof of Proposition~\ref{pro:PhaseTrans}}
\begin{definition}[Duplication Events]\label{defn:DuplicationVar}
Consider any $t\ge0$ and $j\in\cZ_t$. 
Let $B_{t}^{(j)}\in\set{0, 1}$ be the indicator for a duplication occurring by RW~$j\in\cZ_t$ at time $t$. 
\end{definition}
From the W-AC algorithm defined in Algorithm~\ref{alg:WAC}, we have the expression for $B_t^{(j)}$:
\begin{equation}\label{eq:DuplicationVar1}
B_{t}^{(j)} = \setIn{X_{t, j}\neq 0}\setIn{H_{t}^{(X_{t, j}, j)}\ge A}.
\end{equation}
\begin{definition}\label{defn:DeltaCndtion}
Define $\Delta_{t}(A, z)$ as the expected number of newly duplicated RWs at time $t$, given that the current number of active RWs is $Z_t = z$. 
That is, 
\begin{equation}\label{eq:Deltat}
\Delta_{t}(A, z) \triangleq \E\left[ \sum_{j\in\cZ_t} B_{t}^{(j)} \,\middle|\, Z_t = z \right].
\end{equation}
\end{definition}

\subsubsection{Expression for $\Delta_t(A, z)$}
In this subsection, we first derive a closed-form expression for $\Delta_{t}(A, z)$. For simplicity in analysis, conditional on the event $Z_t=z$, we re-label the active RWs in $\cZ_t$ as $[z]$ without loss of generality. That is, we assign each $j\in\cZ_t$ a new index in $[z]$ via a one-to-one mapping. Accordingly, the associated variables $L_t^{(u,j)}$ and $H_t^{(u,j)}$ are re-indexed under this mapping. With a slight abuse of notation, we continue to denote the re-indexed random walk as $j$, where $1\le j\le z$.

According to Definition~\ref{defn:Timing}, each active RW $j\in[z]$ was generated {\it strictly} before time $t$. Consider any $j\in[z]$ and denote $X_{t, j}=u$. A new RW is duplicated from RW $j$ if the following $2$ conditions are satisfied: (i) $u\in\cB$ and (ii) the associated age variable satisfies $H_t^{(u, j)}\ge A$. We denote every initial condition of the active RWs in $[z]$ as:
\begin{align}\label{eq:Izt}
I_{t, z} = \set{(\theta_j, u_j): \theta_j\in[0, t), u_j\in\cB,j\in[z]}.
\end{align}
By the law of total expectation, \eqref{eq:Deltat} can be re-written as
\begin{align}\label{eq:DeltatOri}
\Delta_{t}(A, z) = &\sum_{I_{t, z}}\E\left[\sum_{j \in [z]}B_{t}^{(j)}\,\middle|\, Z_t = z, I_{t, z}\right]\nonumber\\
\times&\Prb\left(I_{t, z}\,\middle|\, Z_t=z\right).
\end{align}

We now focus on analyzing each conditional expectation term $\E\left[B_{t}^{(j)}\,\middle|\, Z_t = z, I_{t, z}\right]$ with $j\in[z]$. 
For each $j\in[z]$, let the lifetime $K^{(j)}$ be defined in Definition~\ref{defn:StoppingTj}. When $t\le K^{(j)}$, we define the first hitting time of node $u\in\cB$ by the RW $j\in[z]$, starting from $u_j\in\cB$, as:
\begin{align}\label{eq:StoppingTju}
K_{u}^{(j)} = \inf\set{\tau\in[\theta_j, t]: X_{\tau, j}=u, X_{\theta_j,j}=u_j}.
\end{align}
For each $j\in[z]$, we have:
\begin{align}\label{eq:DefnB}
&\E\left[B_{t}^{(j)}\,\middle|\, Z_t = z, I_{t, z}\right]=\E\left[B_{t}^{(j)}\,\middle|\, 1\le j\le z, I_{t, z}\right]\nonumber\\
=&\sum_{u\in[N]}\Prb\Big(X_{t, j}=u, H_t^{(u,j)}\ge A\mid 1\le j\le z, I_{t}\Big)\nonumber\\
=&\sum_{u\in[N]}\Prb\Big(X_{t, j}=u, K_u^{(j)}\le t, H_t^{(u,j)}\ge A\mid 1\le j\le z, I_{t}\Big)\nonumber\\
\triangleq&\sum_{u\in[N]}Q_{u,j}.
\end{align}
By the joint distribution factorization, $Q_{u,j}$ defined in \eqref{eq:DefnB} can be computed as follows:
\begin{align}\label{eq:DefnQ}
&Q_{u,j}=\Prb\left(X_{t, j}=u\,\middle|\, 1\le j\le z, I_{t,z}\right)\nonumber\\
&\times\Prb\left(K_u^{(j)}\le t, H_t^{(u,j)}\ge A\,\middle|\, 1\le j\le z, I_{t,z},X_{t, j}=u\right)\nonumber\\
&=\Prb\left(X_{t,j}=u\,\middle|\, 1\le j\le z, I_{t,z}\right)\nonumber\\
&\times\sum_{k=\theta_j}^{t}\Prb\left(K_{u}^{(j)}=k\,\middle|\, 1\le j\le z, I_{t, z}, X_{t, j}=u\right)\nonumber\\
&\times\Prb\left(H_t^{(u,j)}\ge A\,\middle|\,1\le j\le z, I_{t,z}, X_{t,j}=u, K_u^{(j)}=k\right).
\end{align}

Note that RW $j$ was born at time $\theta_j$, and the killing probability $\zeta=1$, it cannot visit the Pac-Man node during the interval $(\theta_j, t]$. Therefore:
\begin{align}\label{eq:Qju1}
Q_{u, j}^{(1)}\triangleq&\Prb(X_{t, j}=u\mid 1\le j\le z, I_{t,z})\nonumber\\
=& \frac{(\frac{N-1}{N})^{t-\theta_j-1}\frac{1}{N}}{(\frac{N-1}{N})^{t-\theta_j}}=\frac{1}{N-1}.
\end{align}

Recall that $k$ is the value of the first hitting time $K_u^{(j)}$. For each $k\in[\theta_j,t]$, the event $K_u^{(j)}=k$ implies that RW $j$ cannot visit node $u$ before time $k$. Therefore:
\begin{align}\label{eq:Qju2}
&Q_{u, j}^{(2,k)}\triangleq\Prb\left(K_u^{(j)}=k\,\middle|\, X_{t, j}=u, 1\le j\le z, I_{t,z}\right)\nonumber\\
&=\frac{\Prb\left(K_u^{(j)}=k, X_{t, j}=u \,\middle|\, 1\le j\le z, I_{t,z}\right)}{\Prb\left(X_{t, j}=u \,\middle|\, 1\le j\le z, I_{t,z}\right)}\nonumber\\
&=\left\{
\begin{aligned}
&\setIn{u_j=u}&& k=\theta_j\\
&\setIn{u_j\neq u}\frac{1}{N-1}\big(\frac{N-2}{N-1}\big)^{k-\theta_j-1}&& \theta_j<k<t\\
&\setIn{u_j\neq u}(\frac{N-2}{N-1})^{t-\theta_j-1}&& k=t.
\end{aligned}
\right.
\end{align}
Consider the condition $H_t^{(u,j)}\ge A$. It implies that RW $j$ cannot visit node $u$ during the interval $[t-A, t-1]$. Therefore,  if $t-A < k$, then $H_t^{(u,j)}\ge A$ and $K_u^{(j)}=k$ are incompatible and cannot occur simultaneously. Therefore:
\begin{align*}
&Q_{u, j}^{(3,k)}\triangleq\Prb\left(H_t^{(u,j)}\ge A \,\middle|\, X_{t, j}=u, K_u^{(j)}=k, I_{t,z}\right)\nonumber\\
&=\frac{\Prb\left(H_t^{(u,j)}\ge A, K_u^{(j)}=k\,\middle|\, X_{t, j}=u, I_{t,z}\right)}{\Prb\left(K_u^{(j)}=k\,\middle|\, X_{t, j}=u, I_{t,z}\right)}.
\end{align*}
We evaluate this quantity in three cases:
\begin{compactenum}[(i)]
\item {\it Case 1}: $k=\theta_j$. Given that $X_{t, j}=u$, by definition of $H_{t}^{(u,j)}$ in Definition~\ref{defn:AgeofVisiting}, we have $H_{t}^{(u,j)}=t - L_{t-1}^{(u, j)}\ge 1>0$. Then:
\begin{align}\label{eq:Qju31}
Q_{u, j}^{(3,k)} =& \setIn{u=u_j}\setIn{t-A\ge \theta_j}\nonumber\\
&\times\left(\setIn{A\ge 1}(\frac{N-2}{N-1})^{A-1}+\setIn{A = 0}\right).
\end{align}
\item {\it Case 2}: $\theta_j<k < t$. Again, since $X_{t, j}=u$, we know $H_{t}^{(u, j)}=t - L_{t-1}^{(u, j)}\ge 1>0$. Then:
\begin{align}\label{eq:Qju32}
Q_{u, j}^{(3,k)}=&\setIn{u\neq u_j}\setIn{t-A\ge k}\nonumber\\
&\times\left(\setIn{A\ge 1}(\frac{N-2}{N-1})^{A-1}+\setIn{A=0}\right).
\end{align}
\item {\it Case 3}: If $k = t$, we have:
\begin{align}\label{eq:Qju33}
Q_{u, j}^{(3,k)}=&\setIn{u\neq u_j}\setIn{A=0}.
\end{align}

\end{compactenum}

Substituting \eqref{eq:Qju1}, \eqref{eq:Qju2}, \eqref{eq:Qju31}, \eqref{eq:Qju32}, and \eqref{eq:Qju33} into \eqref{eq:DefnQ}, substituting  \eqref{eq:DefnQ} into \eqref{eq:DefnB}, and substituting \eqref{eq:DefnB} into \eqref{eq:DeltatOri}, we obtain the expression for $\Delta_{t}(A, z)$ directly as follows:
\begin{align}\label{eq:Deltatnew}
\Delta_{t}(A, z)&= \sum_{I_{t, z}}\Prb\left(I_{t, z}\,\middle|\, Z_t=z\right)\nonumber\\
& \times\sum_{j\in[z]}\left(\sum_{u\in[N]}Q_{u, j}^{(1)}\sum_{k=\theta_j}^{t}Q_{u, j}^{(2,k)}Q_{u, j}^{(3,k)}\right).
\end{align}

\subsubsection{Proof of Part (a)}\label{subsec:PhaseTransSuperMar}

Since each RW hits the Pac-Man at node $1$  with probability $\frac{1}{N}$ at every step, and is killed with probability $\zeta=1$ upon hitting it, then, given the number of active RWs $Z_{t}=z$, the expected number of RWs that die at time $t$ is
\begin{align}\label{eq:DeathRate}
\frac{z}{N}.
\end{align}
From \eqref{eq:Deltat} and \eqref{eq:DeathRate}, given the number of active RWs $Z_t=z$, we derive
\begin{align*}
\E\left[Z_{t+1} \,\middle|\, Z_t=z \right] =& z\Big(1 - \frac{1}{N} + \frac{\Delta_t(A, z)}{z}\Big)\\
&\triangleq z m_t(A, z),
\end{align*}
where we define 
\begin{align}\label{eqn:MultRate}
m_t(A, z)\triangleq 1 -\frac{1}{N} +\frac{\Delta_t(A, z)}{z}.
\end{align}
\begin{lemma}\label{lem:barA}
If there exists a positive integer $\bar{\alpha}$, such that when $A\ge\bar{\alpha}$, the following inequality holds
\begin{align}\label{eq:KeyCondi1}
\sup_{t \ge 0}\sup_{z \in \Z_+}\frac{\Delta_t(A, z)}{z} < \frac{1}{N}.
\end{align}
Then, we have 
\begin{align*}
\Prb\left(\exists\, t_0,\,\forall\, t\ge t_0,\, Z_t=0 \,\middle|\, Z_0=z_0\right)=1
\end{align*}
for all $z_0\ge1$.
\end{lemma}
\begin{IEEEproof}
Let $A\ge\bar{\alpha}$. 
By assumption \eqref{eq:KeyCondi1}, 
\begin{align*}
\frac{\Delta_t(A, z)}{z}<\frac{1}{N}
\end{align*}
holds for all $z\in\Z_+$ and $t\ge 0$. 
Then we have
\begin{align*}
m_t(A, z) = 1 -\frac{1}{N} +\frac{\Delta_t(A, z)}{z} < 1
\end{align*}
holds for all $z\in\Z_+$, and $t\ge 0$. We define
\begin{align*}
\epsilon \triangleq 1 - \sup_{t\ge0, z\in\Z_+} m_t(A, z)>0.
\end{align*}
Therefore, 
\begin{align*}
\E[Z_{t+1}|Z_t=z]=z m_t(A, z) \le (1-\epsilon)z.
\end{align*}
This implies 
\begin{align}\label{eq:SuperMarZ}
\E[Z_{t+1}|Z_t]\le (1-\epsilon)Z_t<Z_t.
\end{align}
It follows that $\{Z_t\}_t$ is a supermartingale. Since $Z_t\geq 0$ for all $t$,  Doob’s Super-Martingale Convergence Theorem ensures that
\begin{align*}
Z_t \to Z_{\ell},\,\,a.s.
\end{align*}
Hence,
\begin{align}\label{eq:ConvergeMean}
\E[Z_t] \to \E[Z_{\ell}].
\end{align}
We now prove that $Z_{\ell}=0$ almost surely by contradiction. If the statement is false, then $\E[Z_\ell]>0$, taking the limit \eqref{eq:ConvergeMean} into \eqref{eq:SuperMarZ}, we obtain
\begin{align*}
\E[Z_{\ell}]\leq (1-\epsilon)\E[Z_{\ell}]<\E[Z_{\ell}],
\end{align*}
which is a contradiction. Thus,  $Z_{\infty}=0$ almost surely. This implies that all RWs are eventually absorbed by the Pac-Man, and hence 
\begin{align*}
\Prb\left(\exists\, t_0,\,\forall\, t\ge t_0,\, Z_t=0 \,\middle|\, Z_0=z_0\right)=1
\end{align*} 
for all $z_0$.
\end{IEEEproof}

In what follows, we prove the existence of $\bar{\alpha}$. To ensure that condition \eqref{eq:KeyCondi1} holds, we start from $Q_{j,u}^{(2,k)}$ defined in \eqref{eq:Qju2} and $Q_{j, u}^{(3, k)}$ defined in \eqref{eq:Qju31}, \eqref{eq:Qju32}, \eqref{eq:Qju33}. Without loss of generality, we consider $A\ge 1$. Specifically: 
\begin{compactenum}[(a)]
\item In \eqref{eq:Qju2}, since $\frac{1}{N-1}<1$ and $A\ge 1$, then 
\begin{align}\label{eq:BoundsQ2}
Q_{j,u}^{(2,k)}<\big(\frac{N-2}{N-1}\big)^{k-\theta_j-1},\,\, \theta_j \le k\le t.
\end{align}
\item In \eqref{eq:Qju31}, \eqref{eq:Qju32} and \eqref{eq:Qju33}, since $\frac{1}{N}\le\frac{N-2}{N}<\frac{N-1}{N}<1$ (when $N\ge 3$) and $A\ge 1$, then
\begin{align}\label{eq:BoundsQ3}
Q_{j, u}^{(3,k)} < (\frac{N-2}{N-1})^{A-1}.
\end{align}
\end{compactenum}
Substituting \eqref{eq:BoundsQ2} and \eqref{eq:BoundsQ3} into \eqref{eq:Deltatnew}, we obtain:
\begin{align}\label{eq:KeyCondi2}
\Delta_t(A,z) <& z\frac{N-1}{N-2}(\frac{N-2}{N-1})^{A-1}\sum_{I_{t,z}}\Prb\left(I_{t,z}\,\middle|\, Z_t=z\right)\nonumber\\
=&z\frac{N-1}{N-2}(\frac{N-2}{N-1})^{A-1}.
\end{align}
To ensure the condition \eqref{eq:KeyCondi1} holds for all $t\ge0$ and $z\in\Z_+$, it suffices to ensure
\begin{align*}
(\frac{N-2}{N-1})^{A-1} < \frac{N-2}{(N-1)N}.
\end{align*}
Since $(\frac{N-2}{N-1})^{A-1}$ is a decreasing function of $A$, then $(\frac{N-2}{N-1})^{A-1} < \frac{N-2}{N(N-1)}$ when $A\ge1+\frac{\ln(\frac{N-2}{(N-1)N})}{\ln(\frac{N-2}{N-1})}\triangleq\bar{\alpha}$. 

\subsubsection{Proof of Part (b)}

Now, we re-visit the recursion:
\begin{equation*}
\E\big[Z_{t+1}\mid Z_t=z\big] = z m_t(A, z),
\end{equation*}
where $m_t(A, z)$ is defined in \eqref{eqn:MultRate}:
\begin{align*}
m_t(A, z)\triangleq 1 -\frac{1}{N} +\frac{\Delta_t(A, z)}{z}.
\end{align*}
\begin{lemma}\label{lem:underlineA}
If there exists a positive integer $\underline{\alpha}$, such that when $A\le\underline{\alpha}$, the following inequality holds:
\begin{align}\label{eq:KeyCondisub1}
\inf_{t\ge0}\inf_{z\in\Z_+}\frac{\Delta_t(A, z)}{z} > \frac{1}{N}.
\end{align}
Then, for any $A\le\underline{\alpha}$, we have 
\begin{align*}
\Prb\left(\exists\, t_0,\,\forall\, t\ge t_0,\, Z_t=0 \,\middle|\, Z_0=z_0\right)<1
\end{align*} 
for all $z_0$.
\end{lemma}
\begin{IEEEproof}
Let $A\le\underline{\alpha}$. 
By assumption \eqref{eq:KeyCondisub1}, 
\begin{align*}
\frac{\Delta_t(A, z)}{z} > \frac{1}{N}
\end{align*}
holds for all $z\in\Z_+$ and $t\ge 0$. 
Then we have
\begin{align*}
m_t(A, z) = 1 -\frac{1}{N} +\frac{\Delta_t(A, z)}{z} > 1
\end{align*}
for all $z\in\Z_+$ and $t\ge 0$. Define
\begin{align*}
\epsilon \triangleq \inf_{t\ge0, z\in \Z_+} m_t(A, z) -1 >0.
\end{align*}
Then, for all $z\in\Z_+$,
\begin{align*}
\E[Z_{t+1}|Z_t=z]=z m_t(A, z) \ge (1+\epsilon)z.
\end{align*}
This implies 
\begin{align}\label{eq:SubMarZ}
\E[Z_{t+1}]\ge (1+\epsilon)\E[Z_t].
\end{align} 
Assume, for contradiction, that 
\begin{align*}
\Prb\left(\exists\, t_0,\,\forall\, t\ge t_0,\, Z_t=0 \,\middle|\, Z_0=z_0\right)=1
\end{align*} 
for any $z_0$. Then, $\lim_{\ell\to\infty}Z_\ell=0$ almost surely, and hence $\lim_{\ell\to\infty}\E[Z_\ell]=0$. However, by recursively applying \eqref{eq:SubMarZ}, we obtain:
\begin{align*}
\lim_{\ell\to\infty}\E[Z_{\ell}] \ge \lim_{\ell\to\infty}(1+\epsilon)^\ell z_0\to\infty,
\end{align*}
which contradicts to assumption. Therefore, 
\begin{align*}
\Prb\left(\exists\, t_0,\,\forall\, t\ge t_0,\, Z_t=0 \,\middle|\, Z_0=z_0\right)<1
\end{align*}
for all $z_0\in\Z_+$. 
\end{IEEEproof}

In what follows, to prove the existence of $\underline{\alpha}$.  
We begin by examining the case where $A=0$. Substituting $A=0$ into \eqref{eq:Deltatnew}, we have:
\begin{align}\label{eq:Deltaz}
\Delta_{t}(0, z)=\frac{N-1}{N} z.
\end{align}
Substituting \eqref{eq:Deltaz} into \eqref{eqn:MultRate}, we observe that the resulting expression satisfies condition \eqref{eq:KeyCondisub1} for all $t\ge0$ and $z\in\Z_+$. This implies that $A=0$ is always a valid threshold under the given condition.

Next, we consider the case $A=1$. From \eqref{eq:Qju1}, \eqref{eq:Qju2}, \eqref{eq:Qju31}$\sim$\eqref{eq:Qju33}, and \eqref{eq:Deltatnew}, we obtain:
\begin{align*}
&\Delta_{t}(1, z) = \sum_{I_{t, z}}\Prb\left(I_{t}\,\middle|\, Z_t=z\right)\nonumber\\
& \times\sum_{j\in[z]}\left(\sum_{u\in\cB}\frac{1}{N}\left(1-\setIn{u\neq u_j}(\frac{N-2}{N-1})^{t-\theta_j-1}\right)\right).
\end{align*}
Since $t-\theta_j>1$, i.e., $t-\theta_j-1\ge0$, and the equality $t=\theta_j+1$ can not hold for every $j\in[z]$ and every $I_{t, z}$, then
\begin{align}\label{eq:Deltatnew1}
\Delta_{t}(1, z)& = \sum_{I_{t, z}}\Prb(I_{t, z}\mid Z_t=z)\nonumber\\
& \times\sum_{j\in[z]}\left(\frac{N-1}{N}-\frac{N-2}{N}(\frac{N-2}{N-1})^{t-\theta_j-1}\right)\nonumber\\
&> \frac{z}{N}\sum_{I_{t, z}}\Prb(I_{t, z}\mid Z_t=z)=\frac{z}{N}.
\end{align}
Substituting \eqref{eq:Deltatnew1} into \eqref{eqn:MultRate}, we see that condition \eqref{eq:KeyCondisub1} is satisfied for all $t\ge0$ and $z\in\Z_+$. This confirms that $A=1$ is always a valid threshold under the given condition.

For any $z_0\in\Z_+$, we choose $\underline{\alpha}=1$, then 
\begin{align*}
\Prb\left(\exists\, t_0,\,\forall\, t\ge t_0,\, Z_t=0 \,\middle|\, Z_0=z_0\right)<1
\end{align*}
for all $A\le\underline{\alpha}$, i.e., $A\in\set{0, 1}$.

\section{Proof of Theorem~\ref{thm:Effectiveness}}\label{Appe:Effectiveness}

We treat a chain of RWs as a {\it single effective} RW and study its limiting behavior, if it exists. Consider an infinite chain of RWs $\set{j_s}_s$. Let $\cA$ be the set of absorbing states, i.e., $\cA=\set{1, w}$ if $\zeta=1$ and $\cA=\set{w}$ if $\zeta\in(0, 1)$.
Let $u\in\cB$ denote the initial location of RW $j_0$. Let $\nu$ be a probability measure on $\cB$, and $\Prb_u$, $\Prb_{\nu}$ be defined in 
Definition~\ref{defn:Initdistribution}. The stopping times of RW $j_0$ with respect to $\cA$, starting from $X_{j_0}(0)=u$ and $X_{j_0}(0)\sim\nu$, are defined as:
\begin{align}
K_{u} \triangleq& \inf\set{t > 0: X_{j_0}(t) \in \cA, X_{j_0}(0) = u},\label{eq:StoppingT}\\
K_{\nu} \triangleq& \inf\set{t > 0: X_{j_0}(t) \in \cA, X_{j_0}(0)\sim\nu}.\label{eq:StoppingTProb}
\end{align}
We have $\Prb_{u}(K_{u}<\infty)=1$ and $\Prb_{\nu}(K_{\nu}<\infty)=1$.

\begin{definition}\label{defn:Activej0}
(Active distribution) Consider a strongly connected graph $\cG$ with absorbing states $\cA$, as defined in Definition~\ref{defn:StrongConnect}. Let a chain of RWs $\set{j_s}_{s\ge0}$ be defined in Definition~\ref{defn:ChainRWs}. Let $K_u$ be defined in \eqref{eq:StoppingT}. For any $t$ and $I\subset\cB$, we define the active distribution of RW $j_0$ at time $t$ as
\begin{align}\label{eq:InitialActiveProb}
\xi_{0; t}(I; u) \triangleq \Prb_u\big(X_{j_0}(t)\in I\mid K_u>t\big).
\end{align}
Let $t_s$ denote the birth time of RW $j_s$, and suppose that its initial location $X_{j_s}(t_s)$ is drawn from a distribution $\nu_s$, which depends on $u$. For any $t\ge t_s$,
we define the active distribution of RW $j_s$ as
\begin{align}\label{eq:ActiveProbs}
\xi_{s; t}(I; u) \triangleq \Prb_{\nu_s}\big(X_{j_s}(t)\in I\mid K_{\nu_s}>t-t_s\big),
\end{align}
where the subscript $\nu_s$ emphasizes that the RW is initialized according to $\nu_s$. The dependence of $\nu_s$ on  $u$ is implicit in this notation $\xi_{s; t}(I; u)$.
\end{definition}
Couple the newly created RW with its parent RW such that, after creation, the new RW is independently reinitialized as an \iid replica of the original RW. That is, it evolves independently and has the same probability distribution as the original RW. Consequently, at any time while the parent RW remains active, the probability distribution of the newly created RW coincides with that of an independent copy of the parent RW. \textit{Recall that we remove the waiting time}, applying this argument recursively, at any time $t$, the active probability distribution of any active RW coincides with that of any ancestor, as long as the ancestor remains active.
Therefore, for any $s > 0$, we have
\begin{align}\label{eq:SameDist}
\xi_{s;t}\overset{d}{=}\xi_{0; t}.
\end{align}
At any time $t$, we re-parameterize the active distribution of the the most recently created (i.e., latest-born) RW as $\xi_t$. Since the chain $\set{j_{s}}_{s\ge0}$ is infinite, we now analyze limiting behavior of the probability distribution $\xi_t$. From \eqref{eq:InitialActiveProb} and \eqref{eq:SameDist}, we have
\begin{align}\label{eq:LatestProb}
\lim_{t\to\infty}\xi_t(I; u) =& \lim_{t\to\infty}\xi_{0; t}(I; u)\nonumber\\ =& \lim_{t\to\infty}\Prb_u\big(X_{j_0}(t)\in I\mid K_u>t\big).
\end{align}
If the limit in \eqref{eq:LatestProb}
exists, it is referred to as the Yaglom limit \cite{foley2017yaglom}. This limit depends on the initial location. Intuitively, the Yaglom limit captures the long-term distribution of the process conditioned on survival. It remains to show that the limit $\lim_{t\to\infty}\xi_t(I;u)$ exists and to derive its explicit expression. 

\begin{definition}\label{defn:DSQ}
(Quasi-Stationary Distribution \cite{collet2012quasi}) 
Consider a strongly connected graph $\cG$ with absorbing states $\cA$, as defined in Definition~\ref{defn:StrongConnect}. Let $K_u$ be defined in \eqref{eq:StoppingT}. We say that $\nu$ is a quasi-stationary distribution (QSD) of RW $j_0$ if, for all $t\ge 0$ and any set $I\subset \cB$, 
\begin{align*}
\nu(I) = \Prb_{\nu}(X_{j_0}(t)\in I\mid K_{u}>t).
\end{align*}
\end{definition}

The following Lemma~\ref{lem:QSDchain} shows that the distribution of a chain of RWs converges asymptotically to that of a single RW conditioned on long-term survival. This provides a way to obtain the explicit expression of $\lim_{t\to\infty}\xi_t(I;u)$.

\begin{lemma}\label{lem:QSDchain}
Consider a robustly connected graph $\cG$ with absorbing states $\cA$, as defined in Definition~\ref{defn:StrongConnect}. Let $\set{j_{s}}_{s\ge0}$ be an infinite chain, as defined in Definition~\ref{defn:ChainRWs}. Suppose the initial RW $j_0$ starts at node $u\in\cB$.  We define the distribution of the chain at time $t$ as
\begin{align}\label{eq:ChainDist}
\pi_{\text{chain},t} \triangleq \xi_t.
\end{align} 
Let $t\to\infty$, the distribution of a chain is convergent:
\begin{align}\label{eq:QSD}
\lim_{t\to\infty}\pi_{\text{chain}, t} = \nu^{(\zeta)}
\end{align}
where $\nu^{(\zeta)}$ is the  left normalized leading eigenvector of $Q^{(\zeta)}$ \big(as defined in \eqref{eq:NewTransMatCase1} and \eqref{eq:NewTransMatCase2}\big).
\end{lemma}
\begin{IEEEproof}
From Definition~\ref{defn:StrongConnect}, the submatrix $Q^{(\zeta)}$ is irreducible. By \cite[Section~2]{foley2017yaglom} or \cite[Theorem~16.11]{billingsley1995probability}, the irreducibility of $Q^{(\zeta)}$ ensures the existence of the corresponding Yaglom limits \big(see \eqref{eq:LatestProb}\big), which is convergent in total variation.

Moreover, since $Q^{(\zeta)}$ is irreducible and aperiodic, any existing Yaglom limit (with any initial state $u$)
coincides with a QSD, as established in \cite[Proposition~1]{foley2017yaglom}. Therefore, the Yaglom limit in \eqref{eq:LatestProb} is a QSD for every  $u\in\cB$.

In our case, each initial RW is defined on a finite state space $\cV$ with a nonempty absorbing set $\cA$. The restricted transition matrix $Q^{(\zeta)}$ on the transient states $\cB$ is reducible and aperiodic. According to \cite{darroch1965quasi}, the QSD exists and is unique. As a result, the Yaglom limit in \eqref{eq:LatestProb} converges to the same QSD for all initial states $u\in\cB$.  

Meanwhile, the QSD can be calculated as the leading left eigenvector of $Q^{(\zeta)}$, normalized to sum to one \cite[Eqn.~(10) and the third equation on p.~99]{darroch1965quasi}. Thus, according to \eqref{eq:ChainDist}, the limiting distribution $\lim_{t\to\infty}\pi_{\text{chain}, t}$ is given by: 
\begin{align*}
\lim_{t\to\infty}\pi_{\text{chain}, t}= \nu^{(\zeta)}.
\end{align*}
\end{IEEEproof}

In each chain of RWs, every child inherits the current model state (i.e., ${\bf x}_t$) from its parent. As a result, under the RW-SGD algorithm, each infinite chain asymptotically behaves as if a single effective RW is solving a surrogate optimization problem with a time-varying sampling distribution $\tilde{\pi}_t$. Specifically: 
\begin{compactenum}
\item When $\zeta=1$, the absorbing state $\cA=\set{1, w}$, so $\tilde{\pi}_t = [0, \pi_{\text{chain},t}]$, where $\pi_{\text{chain};t}$ is a discrete distribution supported on a finite set of size $N$, and
\begin{align*}
\lim_{t\to\infty}\tilde{\pi}_t=[0, \nu^{(1)}].
\end{align*}
\item When $0<\zeta<1$, the absorbing state $\cA=\set{w}$, so $\tilde{\pi}_t = \pi_{\text{chain},t}$, where $\pi_{\text{chain};t}$ is a discrete distribution supported on a finite set of size $N+1$, and 
\begin{align*}
\lim_{t\to\infty}\tilde{\pi}_t= \nu^{(\zeta)}.
\end{align*}
\end{compactenum}

\section{The effective transition probability matrix $P^{(\zeta)}_{\text{chain}}$}\label{Appe:UniqueP}
In fact, as discussed before, we condense the time interval between the termination of the last RW and the creation of the next RW,
where $\cA$ represents the set of absorbing states (as defined in Appendix~\ref{Appe:Effectiveness}), i.e., $\cA=\set{1, w}$ if $\zeta=1$ and $\cA=\set{w}$ if $\zeta\in(0, 1)$.
For any nodes $u, v$, the transition probability matrix $P_\text{chain}$ can be written as
\begin{align*}
\left[P_\text{chain}\right]_{uv} = \Prb\left(\set{X_j(1)=v}\mid \set{X_j(0)=u, X_j(1)\notin\cA}\right).
\end{align*}
Taking marginal distribution of the chain at time $0$ as $\mu$ and applying Bayes’ rule, we obtain:
\begin{align*}
\left[P_\text{chain}\right]_{uv} =& \frac{\Prb\left(\set{X_j(1)=v, X_j(0)=u, X_j(1)\notin\cA}\right)}{\Prb\left(\set{X_j(0)=u, X_j(1)\notin\cA}\right)}\nonumber\\
=& \frac{\mu_uQ^{(\zeta)}_{uv}}{\mu_u\sum_{v}Q^{(\zeta)}_{uv}} =\frac{Q^{(\zeta)}_{uv}}{\sum_{v}Q^{(\zeta)}_{uv}}.
\end{align*}

\section{Proof of Proposition~\ref{pro:Bounds}}\label{Appe:Bounds}

\noindent\textit{Proof of Part (1)}.

Since a chain of RWs behaves like a single RW that never dies, we can apply the convergence results of RW-SGD. According to \cite{10.5555/3618408.3618786, doi:10.1137/08073038X, 10.5555/3327546.3327656}, the standard RW-SGD algorithm converges to a deterministic limit when the stepsize $\eta_t$ decreases with the number of iterations and tends to $0$. Consequently, under the same stepsize condition, a chain of RWs converges to the optimizer of the surrogate optimization problem \eqref{eq:SurOpt}.

Let $\tilde{\bf x}^*$ be the optimizer of either \eqref{eq:SurOpt}. Applying strong convexity, we obtain:
\begin{align*}
f({\bf x}^*)\geq f(\tilde{\bf x}^*) + \langle\nabla f(\tilde{\bf x}^*), {\bf x}^* - \tilde{\bf x}^*\rangle + \frac{\mu}{2}\|{\bf x}^* - \tilde{\bf x}^*\|^2.
\end{align*}
Exchanging ${\bf x}^*$ and $\tilde{\bf x}^*$, we have:
\begin{align*}
f(\tilde{\bf x}^*)\geq f({\bf x}^*) + \langle\nabla f({\bf x}^*), \tilde{\bf x}^* - {\bf x}^*\rangle + \frac{\mu}{2}\|\tilde{\bf x}^* - {\bf x}^*\|^2.
\end{align*}
It follows that
\begin{align*}
\langle \nabla f(\tilde{\bf x}^*)-\nabla f({\bf x}^*), \tilde{\bf x}^*-{\bf x}^*\rangle\ge\mu\|\tilde{\bf x}^*-{\bf x}^*\|^2
\end{align*}
Note that $\nabla f({\bf x}^*)=0$, by Cauchy–Schwarz inequality,
\begin{align*}
\mu\|{\bf x}^* - \tilde{\bf x}^*\|^2\le  \|\nabla f(\tilde{\bf x}^*)\|\|{\bf x}^* - \tilde{\bf x}^*\|.
\end{align*}
Therefore
\begin{align*}
\|\tilde{\bf x}^\star - {\bf x}^\star\|\leq\frac{1}{\mu}\|\nabla f(\tilde{\bf x}^\star)\|.
\end{align*}
The equality holds when $\nabla f(\tilde{\bf x}^\star)$ is co-linear with $\tilde{\bf x}^\star - {\bf x}^\star$ \cite{boyd2004convex}.

Similarly, using the $L$-Lipschitz condition, we derive:
\begin{align*}
L\|{\bf x}^* - \tilde{\bf x}^*\|\geq \|\nabla f({\bf x}^*) - \nabla f(\tilde{\bf x}^*)\|.
\end{align*}
By the optimality conditions, we have:
\begin{align*}
\nabla f({\bf x}^*) = 0,
\end{align*}
which implies
\begin{align*}
\|{\bf x}^* - \tilde{\bf x}^*\|\geq \frac{1}{L}\|\nabla f(\tilde{\bf x}^*)\|.
\end{align*}
The equality holds when $\tilde{\bf x}^\star - {\bf x}^\star$ aligns with the directional of maximal curvature of $f({\bf x})$ \cite{boyd2004convex}.

\noindent\textit{Proof of Part (2)}.
This proof follows the same argument as in \cite[Theorem~1]{10619692}, 
with the necessary substitutions under our setting. 
Specifically, by Theorem~\ref{thm:Effectiveness}, the modified stationary distribution of a single effective RW (i.e., the chain of RWs) 
is $\tilde{\nu}^{(\zeta)}$ with $\zeta\in(0,1]$. 
The corresponding transition probability matrix is given in Appendix~\ref{Appe:UniqueP}. 
Based on Assumption~\ref{assu:Boundedgrad}, we set $w(u)=1$ for all $u\in\cV$.
By substituting $\tilde{\nu}^{(\zeta)}$, $P_{\text{chain}}$, $\gamma_{\text{chain}}$, and the original sampling distribution $\pi$ into the proofs of \cite[Lemmas~1, 2, Theorem~1]{10619692}, and by \textit{artificially condensing the time interval between the child and its parent}, we obtain the following bounds:
\begin{align*}
\E\|\tilde{\bf x}_T - {\bf x}^\star\| \le &\, 
2(1-\eta\mu)^T\|{\bf x}_0 - {\bf x}^\star\|^2 
+ \frac{\eta L\sigma^2}{\gamma_{\text{chain}}\mu^2}\\
+& \frac{\|\tilde{\nu}^{(\zeta)}-\pi\|_{\text{TV}}^2\sigma^2L}{\mu^3}.
\end{align*}

\section{Additional Simulations}\label{Appe: simulations}
In this section, we present additional simulation results of Section~\ref{sec: simulations}. 

\subsection{Additional Graphs}\label{subsec:AdditionalGraphs}
Besides the complete graph, we also consider three other connected graphs, including: (i) a random regular graph with degree $d=8$ (expander graph \cite{expanderdegree}), (ii) a regular graph with degree $d=2$ (ring topology), and (iii) an \ER graph with the edge probability $p=0.1$.

\subsection{Data Partitioning for the Public Benchmark Dataset}\label{subsec:partitioning}

 We use the standard MNIST handwritten digit dataset \cite{deng2012mnist}. The dataset is evenly partitioned into $100$ disjoint subsets, with each node is assigned a unique subset. We consider both \iid and non-\iid data partitioning schemes for distributing data across network nodes. In the \iid case, the dataset is uniformly and independently divided into $100$ disjoint subsets, one for each node. In the non-\iid case, data heterogeneity is introduced by sampling from a Dirichlet distribution \cite{hsu2019measuring}: The concentration parameter $\alpha$ controls the degree of heterogeneity: as $\alpha \to 10$, the partitioning approaches the \iid case; as $\alpha \to 0$, the data becomes highly non-\iid Throughout this paper, we set $\alpha = 0.5$, which corresponds to a moderate level of non-\iid partitioning. Under both partitioning schemes, whenever a RW visits a node, it uniformly samples a mini-batch of size $B=256$ from the node's local data to perform a training update. 

\subsection{Comparison with \textsc{Gossip-based SGD}}\label{subsec:Gossip}
Another baseline we consider is the classical \textsc{Gossip-based SGD} \cite{pmlr-v119-koloskova20a}. In this scheme, each node transmits its locally updated model to all of its neighbors at every iteration, and model parameters are updated via neighborhood averaging. We incorporate the same adversarial setting as before: the Pac-Man node independently terminates each incoming model update with the termination probability $\zeta$. As a result, the Pac-Man node is unable to reliably incorporate all information from its neighbors, leading to biased and incomplete aggregation.

\begin{figure}[htbp]
\centering
\includegraphics[width=0.7\linewidth]{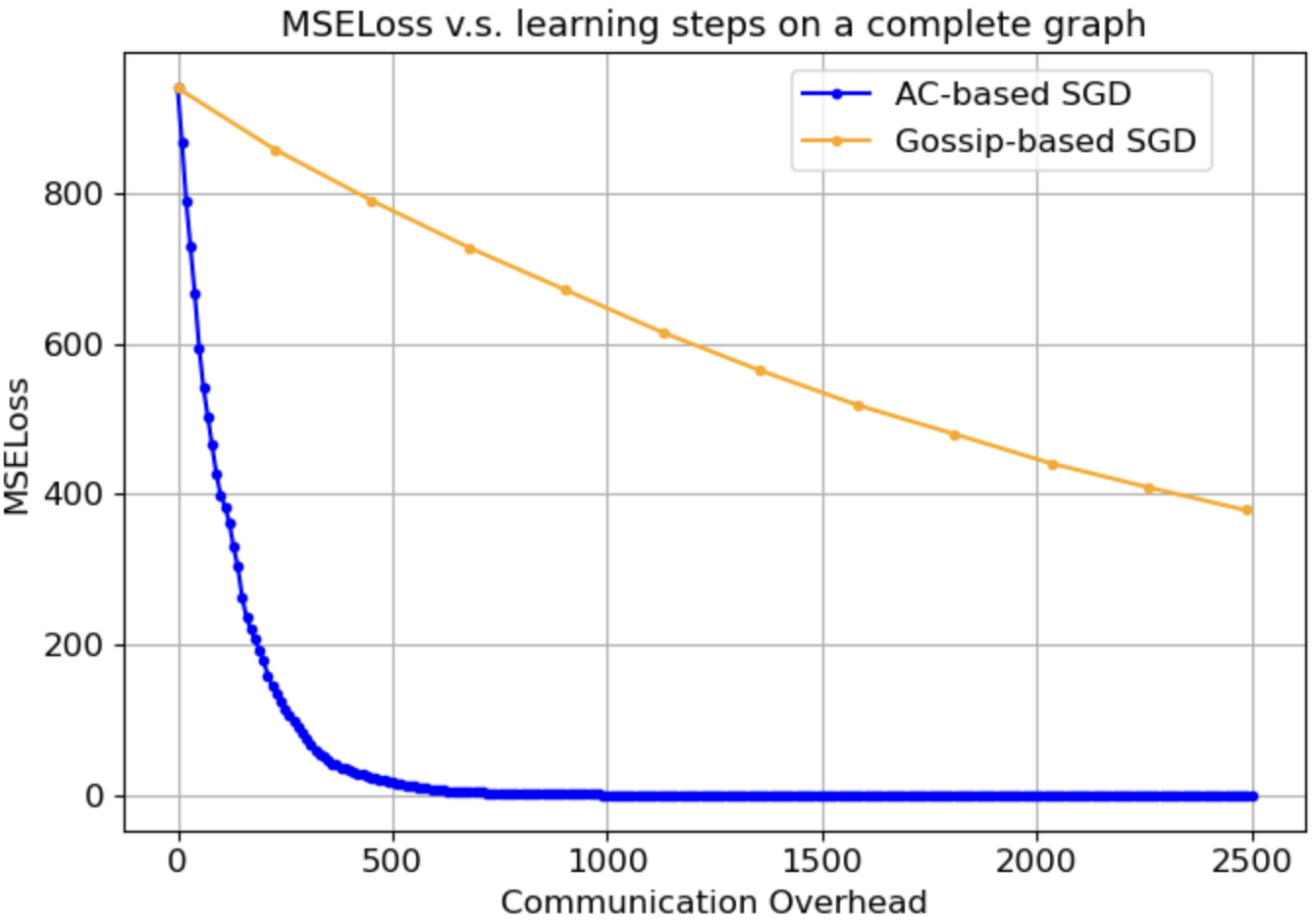}
\caption{Loss v.s. learning steps on a complete graph: comparison between \ac and gossip-based SGD.}
\label{fig:zetaregular}
\end{figure}

Fig.~\ref{fig:zetaregular} shows that \textsc{AC-based SGD} algorithm converges significantly faster than \textsc{Gossip-based SGD}: the loss curve of \textsc{AC-based SGD} (blue curve) exhibits much faster decay than that of \textsc{Gossip-based SGD} (orange curve). In this experiment, we set $\zeta=0.8<1$. This behavior is expected. For the \textsc{AC-based SGD}, we compute the average number of RWs at each time step, and use this quantity as the number of communications per time step. For the \textsc{Gossip-based SGD}, the number of communications at each time step is $2N$. In Fig.~\ref{fig:zetaregular}, each tick on the horizontal axis represents the average number of RWs per time step under \textsc{AC-based SGD}. \textsc{Gossip-based SGD} propagates information through repeated local averaging, which results in diffusive information spread and systematic attenuation of gradient updates. Consequently, it incurs substantial communication overhead. In contrast, \textsc{AC-based SGD} preserves the strength of informative gradients and enables faster (in terms of communication overhead) global impact.

\subsection{Boundedness and Persistence}\label{subsec: Boundedness}

\begin{figure}[htbp]
  \centering
  \begin{minipage}[b]{0.45\linewidth}
    \centering
    \includegraphics[width=\linewidth]{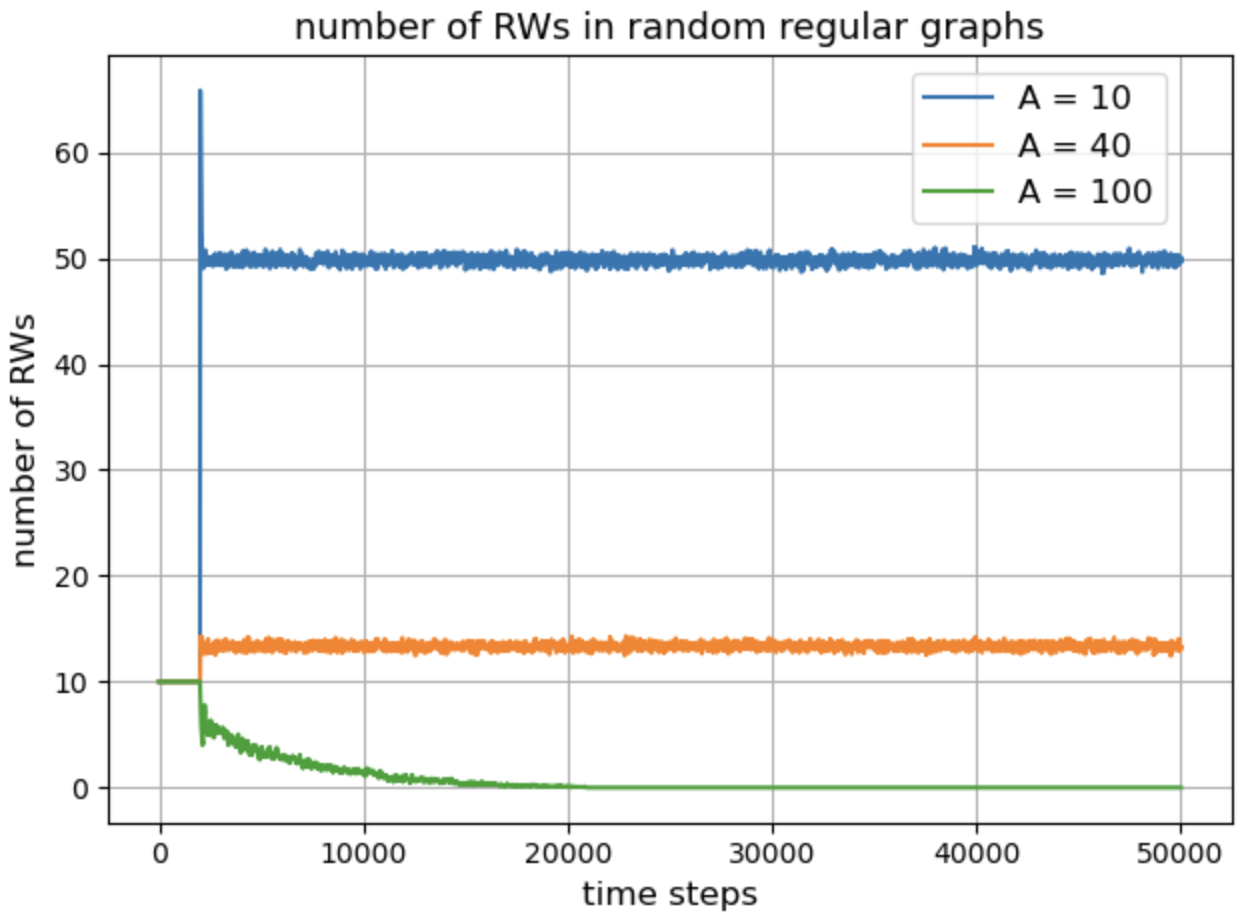}
    \caption*{(a) random regular graph}
  \end{minipage}
  \hfill
  \begin{minipage}[b]{0.45\linewidth}
    \centering
    \includegraphics[width=\linewidth]{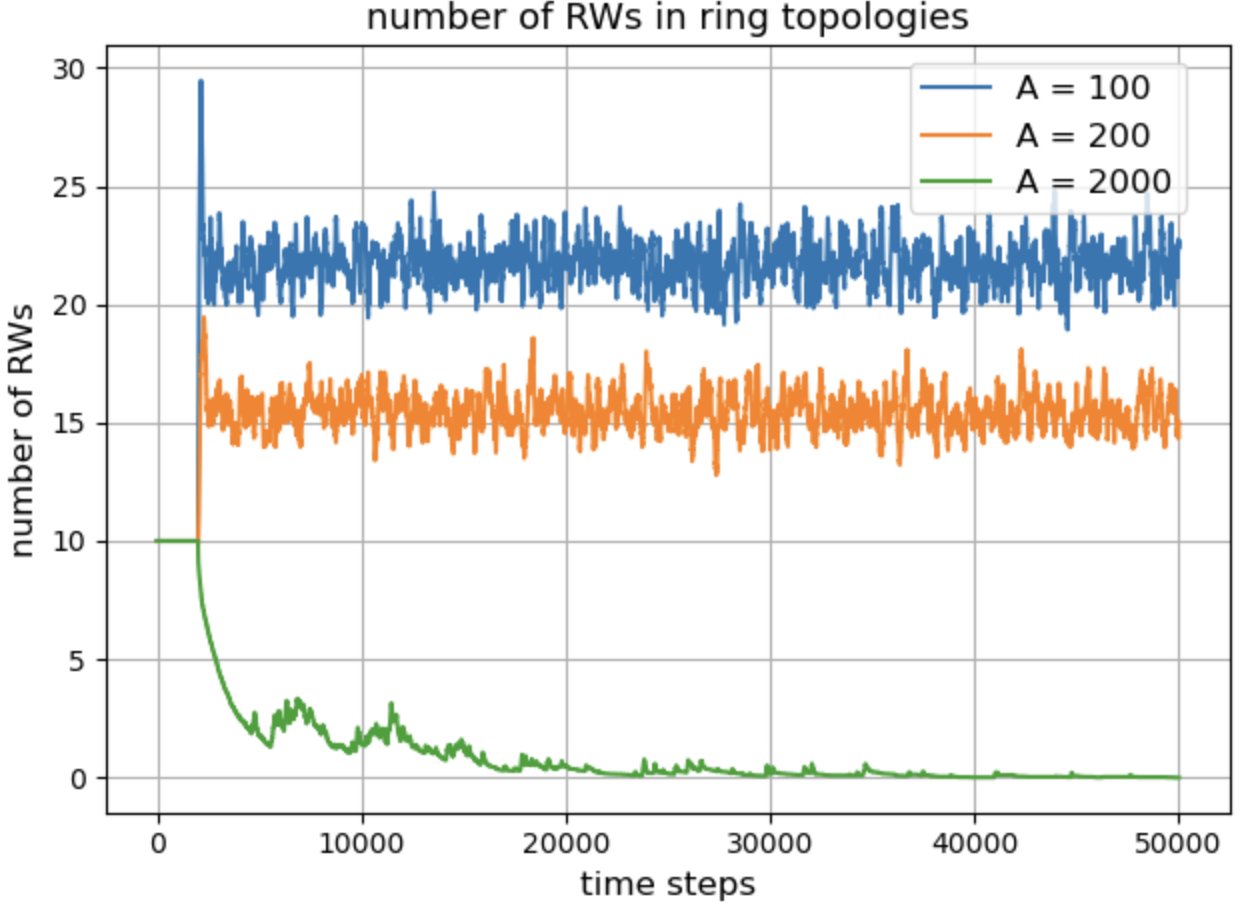}
    \caption*{(b) ring topology}
  \end{minipage}
    \hfill
  \begin{minipage}[b]{0.45\linewidth}
    \centering
    \includegraphics[width=\linewidth]{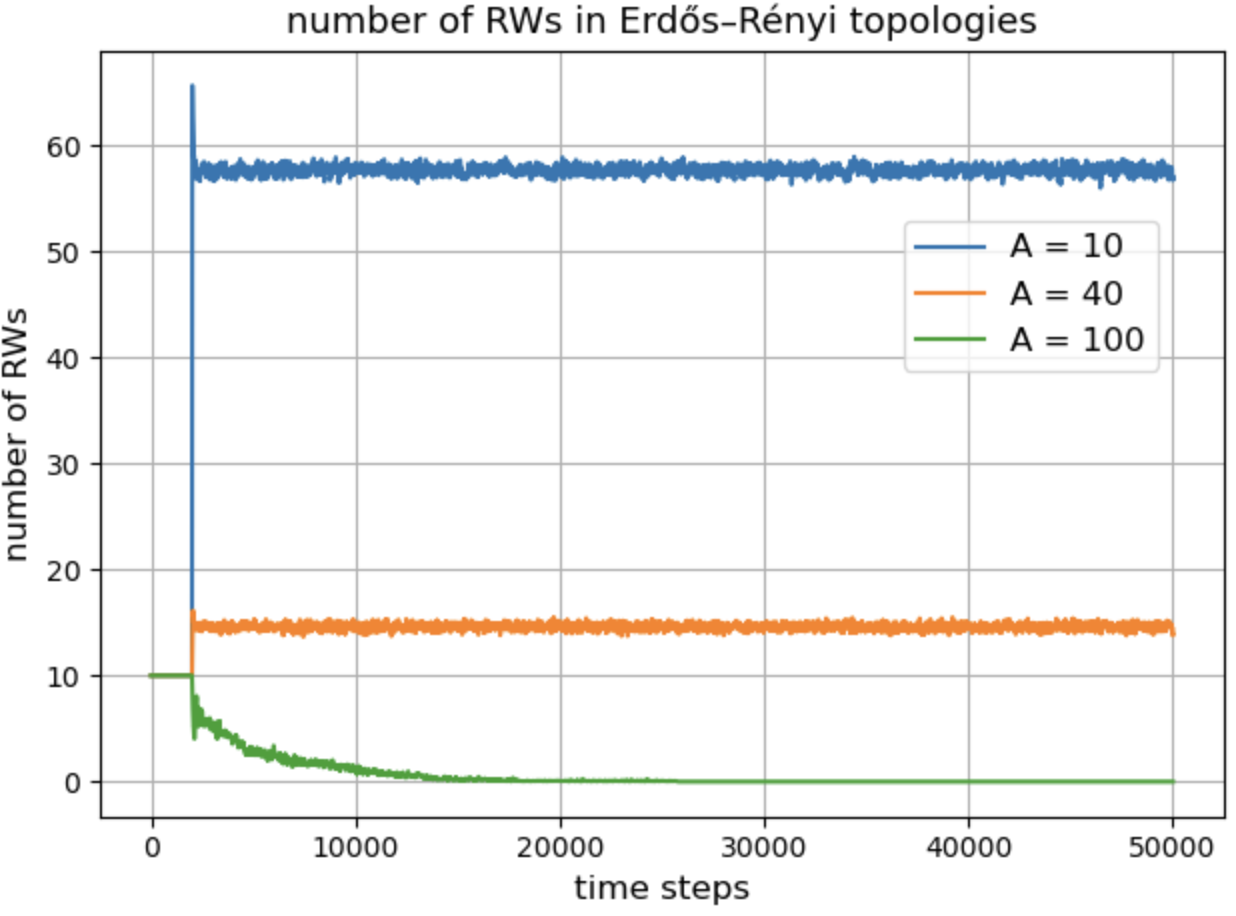}
    \caption*{(c) \ER graph}
  \end{minipage}
  \caption{Number of RWs over time on different graphs.}
  \label{fig: boundedness}
\end{figure}

In Fig.~\ref{fig: boundedness}, we plot the RW population process $\set{Z_t}_{t\in\Z}$ over time on reglar, ring, and \ER graphs. The $y$-axis indicates the number of RWs (averaged over 100 iterations), and the $x$-axis denotes the time steps. On each graph, when the threshold $A$ is small (note that the threshold for ``small'' and ``large'' varies by graph), the RW population  stabilizes at a relatively large value. In contrast, with  large $A$, RW duplication becomes relatively rare, leading to population extinction. 

\begin{figure}[h]
\centering
  \centering
  \begin{minipage}[b]{0.45\linewidth}
    \centering
    \includegraphics[width=\linewidth, height=0.9\linewidth]{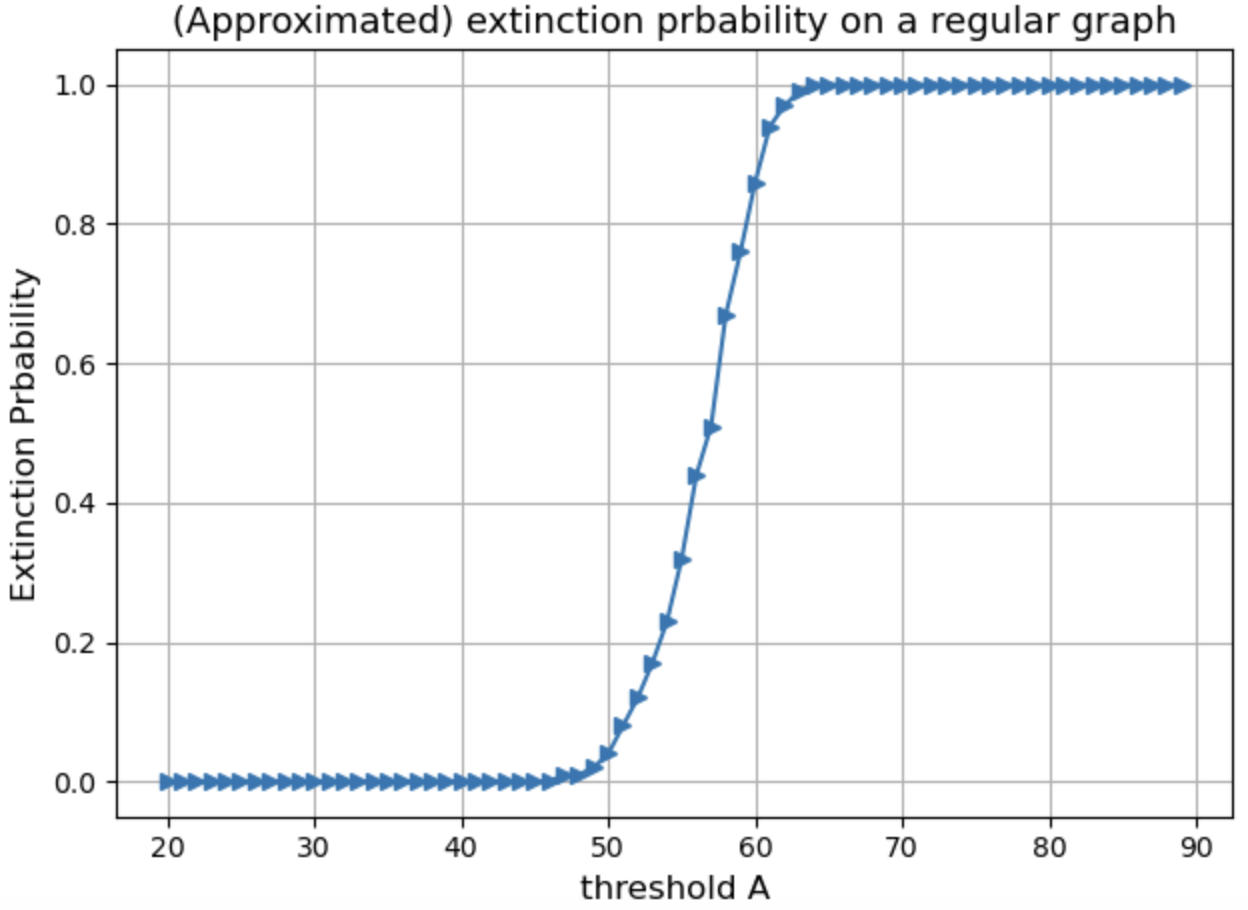}
    \caption*{(a) random regular graph}
  \end{minipage}
  \hfill
  \begin{minipage}[b]{0.45\linewidth}
    \centering
    \includegraphics[width=\linewidth, height=0.9\linewidth]{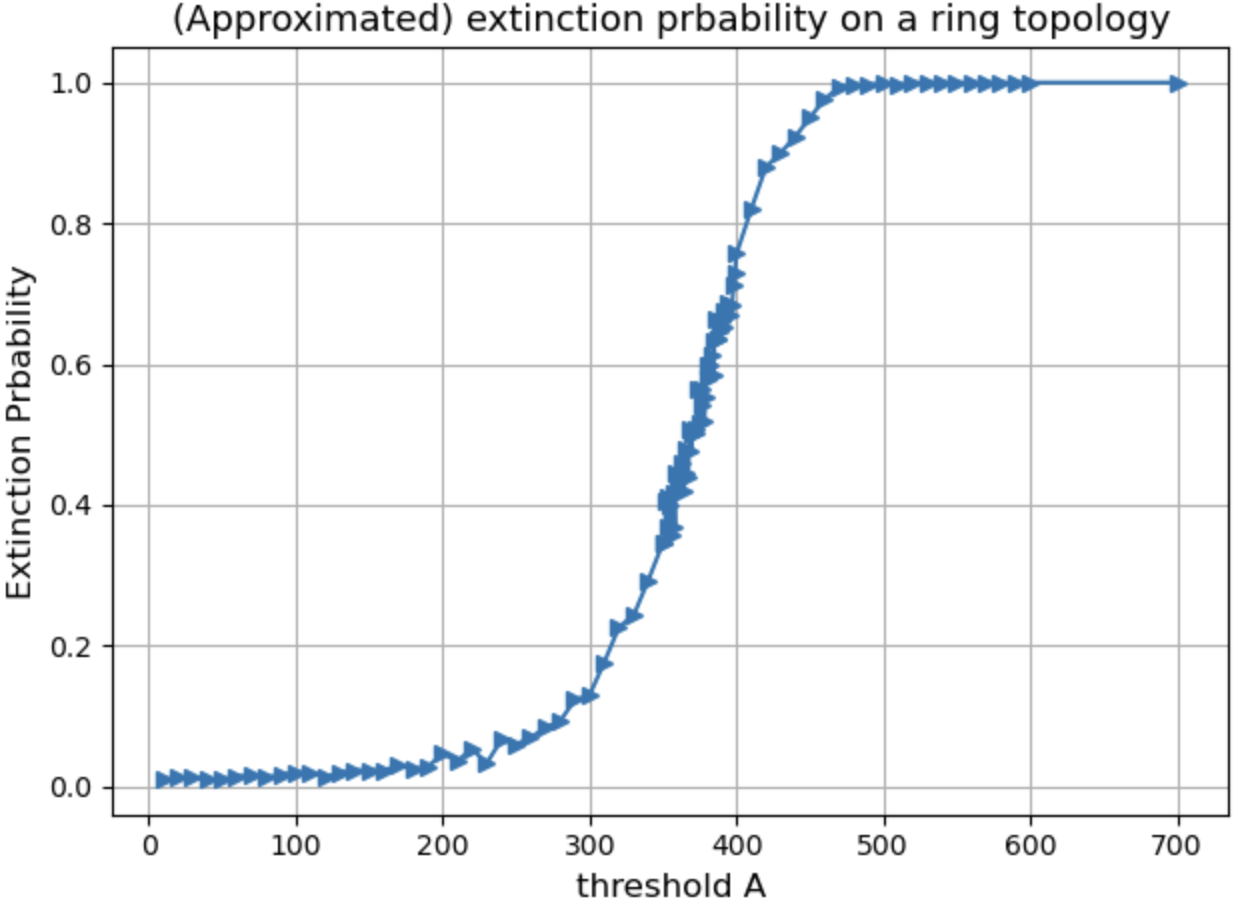}
    \caption*{(b) ring topology}
  \end{minipage}
  \hfill
  \begin{minipage}[b]{0.45\linewidth}
    \centering
    \includegraphics[width=\linewidth, height=0.9\linewidth]{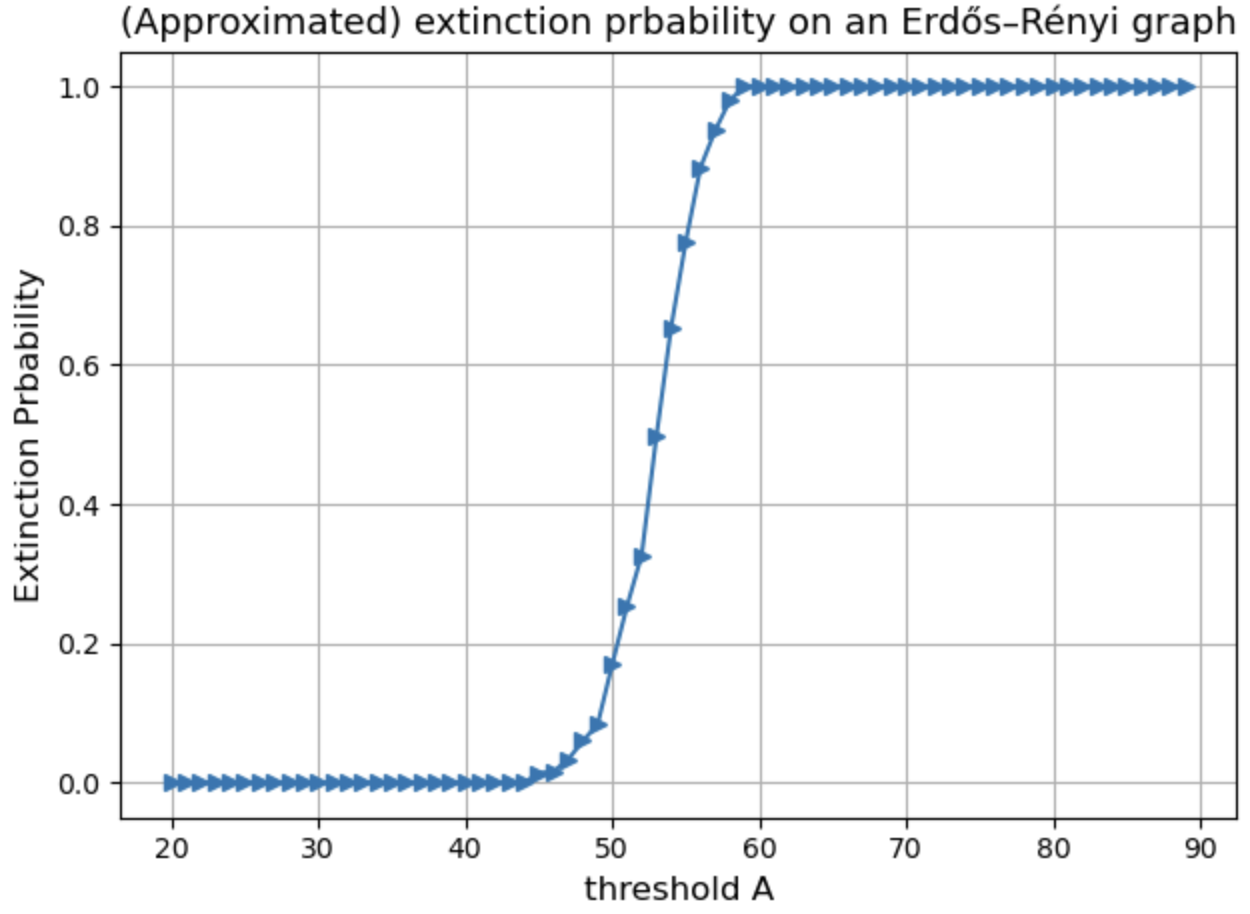}
    \caption*{(c) \ER graph}
  \end{minipage}
\caption{Extinction probability vs. threshold $A$ under different graphs.}
\label{fig: extinction more}
\end{figure}

Although Proposition~\ref{pro:PhaseTrans} focuses exclusively on a simplified version of the AC algorithm on a complete graph, we extend our investigation by simulating the \ac algorithm extinction behavior on regular, ring and Erd\H{o}s--R\'enyi graphs. In Fig.~\ref{fig: extinction more}, the $y$-axis shows the approximated extinction probability\footnote{We approximate the extinction probability by running a large number of simulations over a long time horizon and computing the ratio of runs in which the RW population goes extinct to the total number of runs.}, while the $x$-axis denotes the value of $A$. when $A$ exceeds a critical value (which depends on the graph topology), extinction occurs with probability $1$. Conversely, when $A$ falls below this critical threshold, the extinction probability drops sharply and approaches zero for sufficiently small values of $A$. 

\subsection{Convergence}\label{subsec: convergence}

\begin{figure}[h]
\centering
  \centering
  \begin{minipage}[b]{0.45\linewidth}
    \centering
    \includegraphics[width=\linewidth, height=0.9\linewidth]{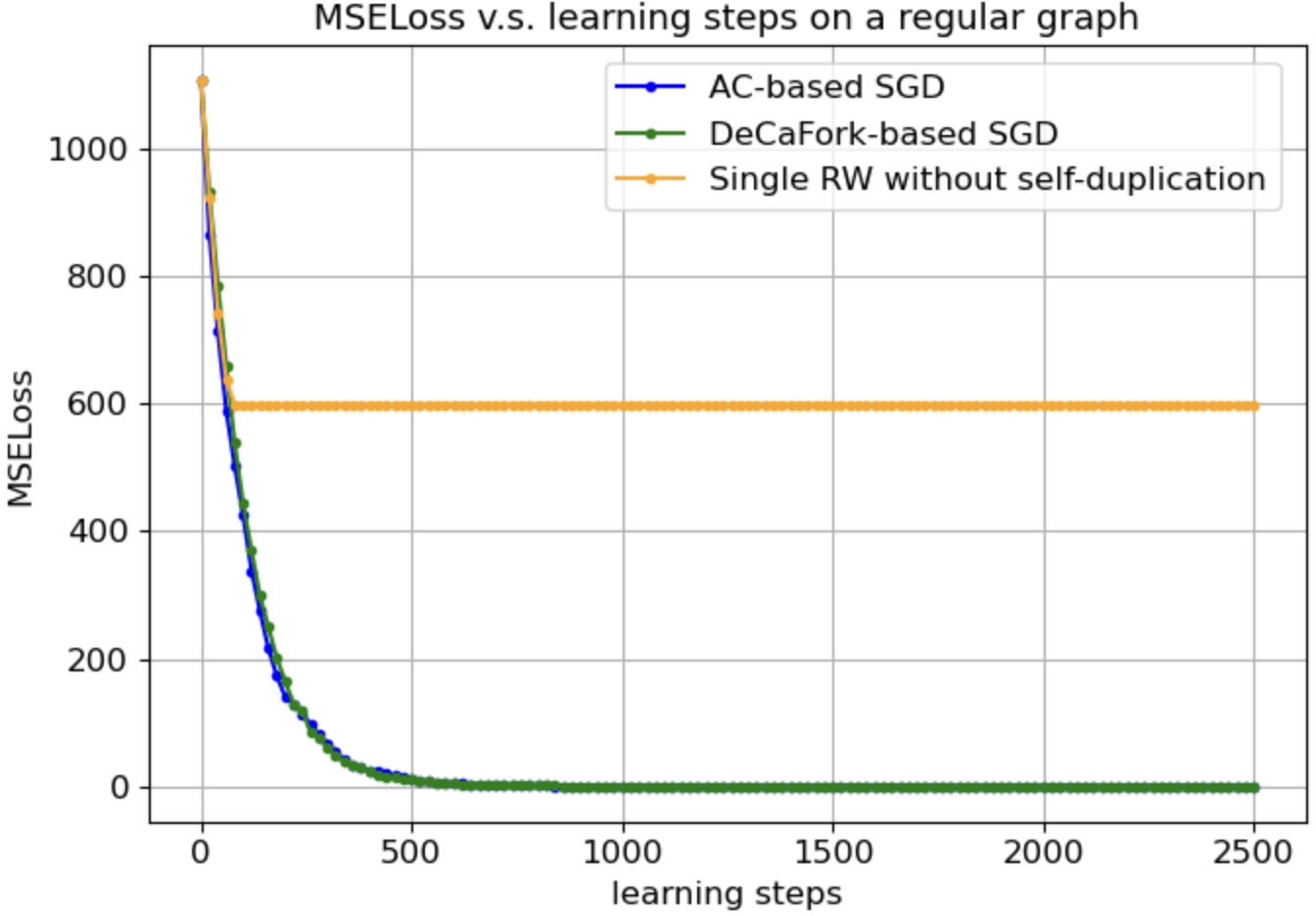}
    \caption*{(a) random regular graph}
  \end{minipage}
  \hfill
  \begin{minipage}[b]{0.45\linewidth}
    \centering
    \includegraphics[width=\linewidth, height=0.9\linewidth]{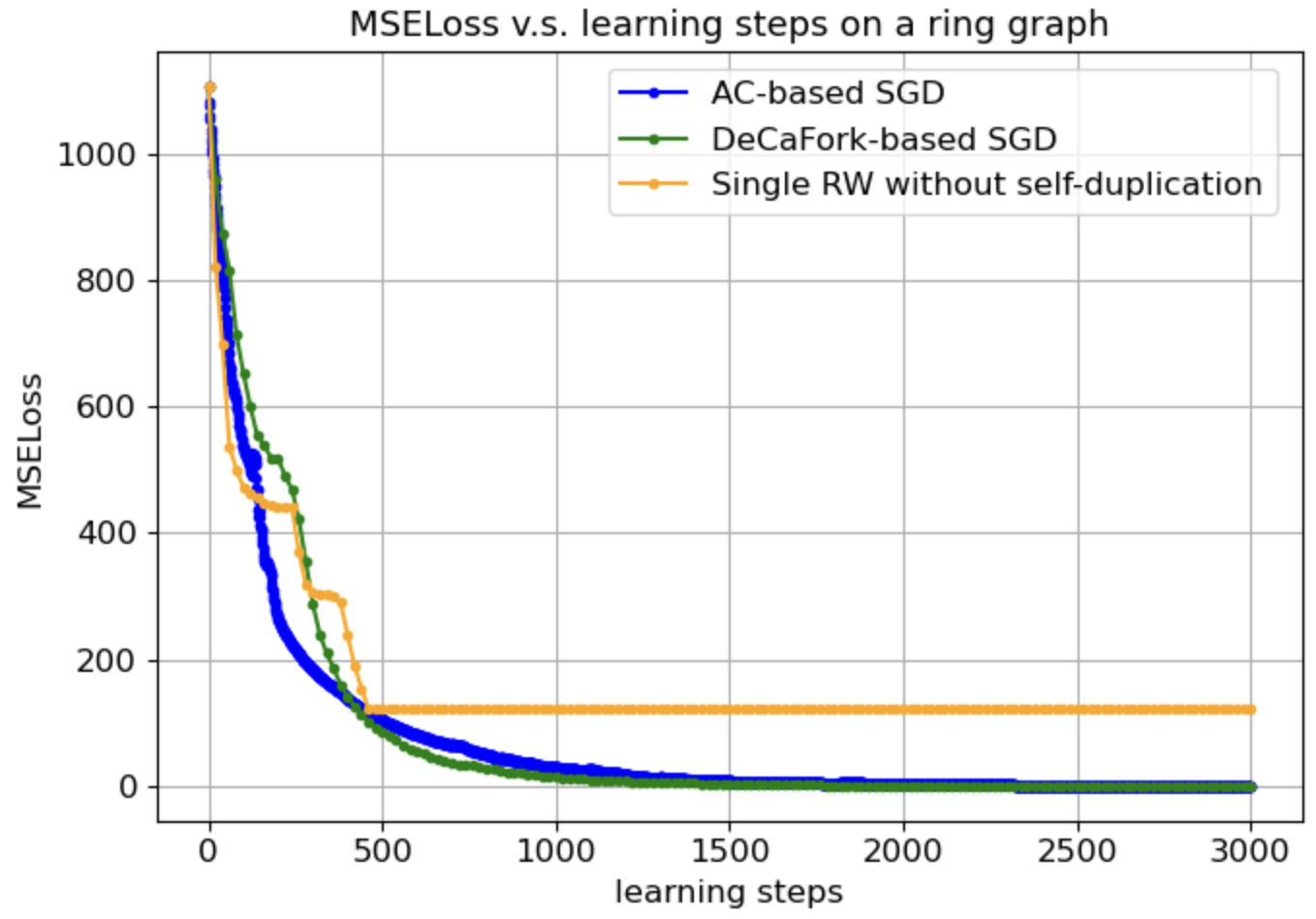}
    \caption*{(b) ring topology}
  \end{minipage}
  \hfill
  \begin{minipage}[b]{0.45\linewidth}
    \centering
    \includegraphics[width=\linewidth, height=0.9\linewidth]{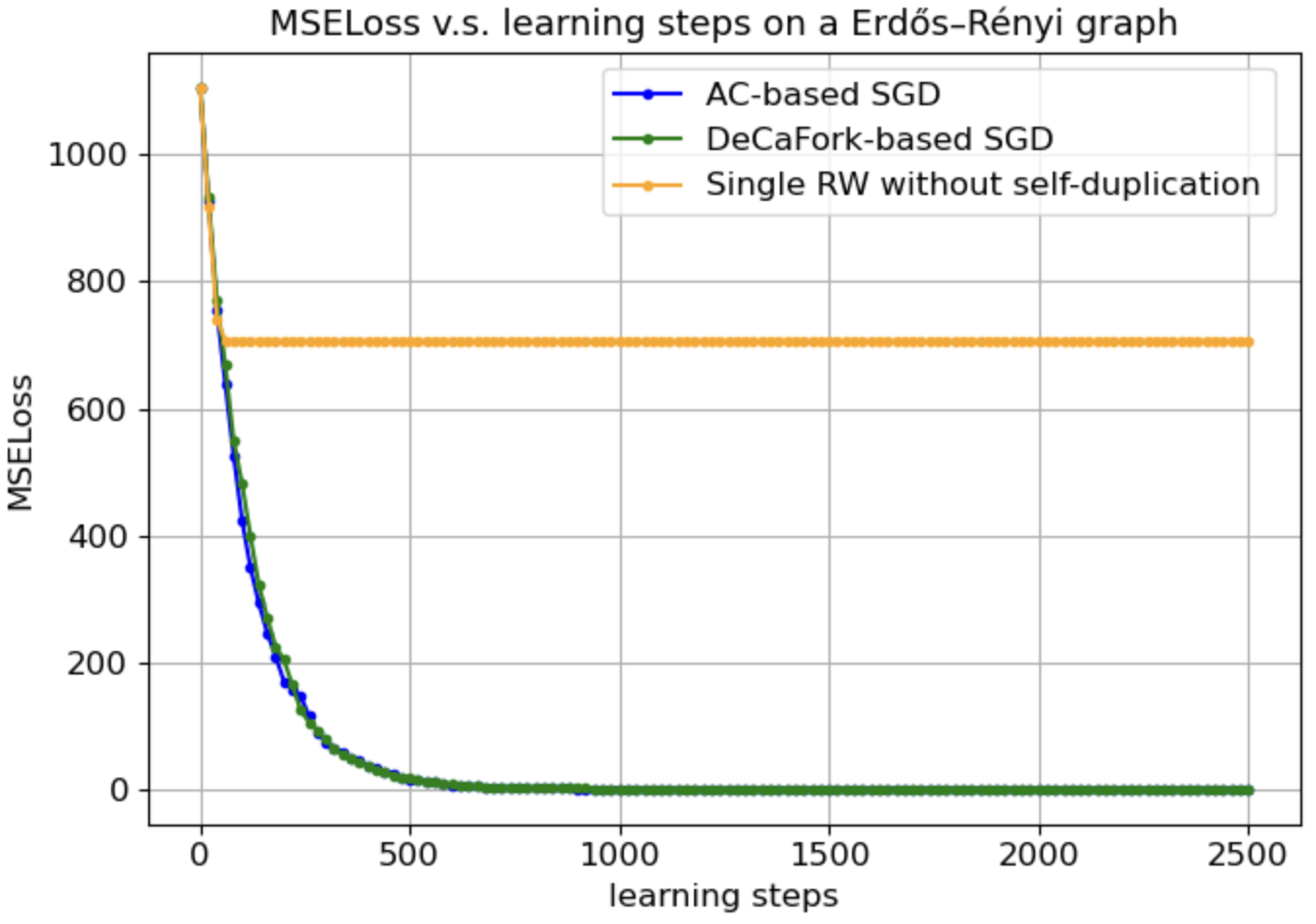}
    \caption*{(c) Erd\H{o}s--R\'enyi graph}
  \end{minipage}
\caption{Loss function v.s. learning steps on different graphs.}
\label{fig:Convergence}
\end{figure}

We begin with experiments on the synthetic dataset. Fig.~\ref{fig:Convergence} presents the the convergence performance of RW-SGD under the proposed \ac algorithm, in comparison with the \DeCa baseline, on regular, ring, and Erd\H{o}s--R\'enyi graphs.  In each subfigure, the $y$-axis represents the value of the global loss, while the $x$-axis denotes the number of time steps. Across different graph topologies, the loss consistently decreases over time and eventually reaches zero, indicating effective convergence.

The distance between the new convergence point and the original optimal solution, along with the corresponding theoretical bounds in the synthetic dateset, is reported in Table~\ref{tab:Bounds}. This table empirically validate the theoretical guarantees established in Proposition~\ref{pro:Bounds}.

\begin{table}[h]
\centering
\begin{tabular}{|c|c|c|c|c|}
\hline
graph type& Complete & Regular & Ring & Erd\H{o}s--R\'enyi \\
\hline
$\|\tilde{\bf x}^* - {\bf x}^*\|$ & $0.019$ & $0.015$ & $0.051$ & $0.018$\\
\hline
$\frac{1}{L}\|\nabla f(\tilde{\bf x}^\star)\|$ & $0.015$ & $0.010$  & $0.048$ & $0.013$\\
\hline
$\frac{1}{\mu}\|\nabla f(\tilde{\bf x}^\star)\|$ & $0.044$ & $0.032$ & $0.146$ & $0.035$\\
\hline
\end{tabular}
\caption{$\|\tilde{\bf x}^* - {\bf x}^*\|$ and its bounds on different graphs.}
\label{tab:Bounds}
\end{table}

We then conducted experiments on the real-world dataset, considering both \iid partitioning and non-\iid partitioning scenarios.

\begin{figure}[h]
\centering
  \centering
  \begin{minipage}[b]{0.45\linewidth}
    \centering
    \includegraphics[width=\linewidth, height=0.9\linewidth]{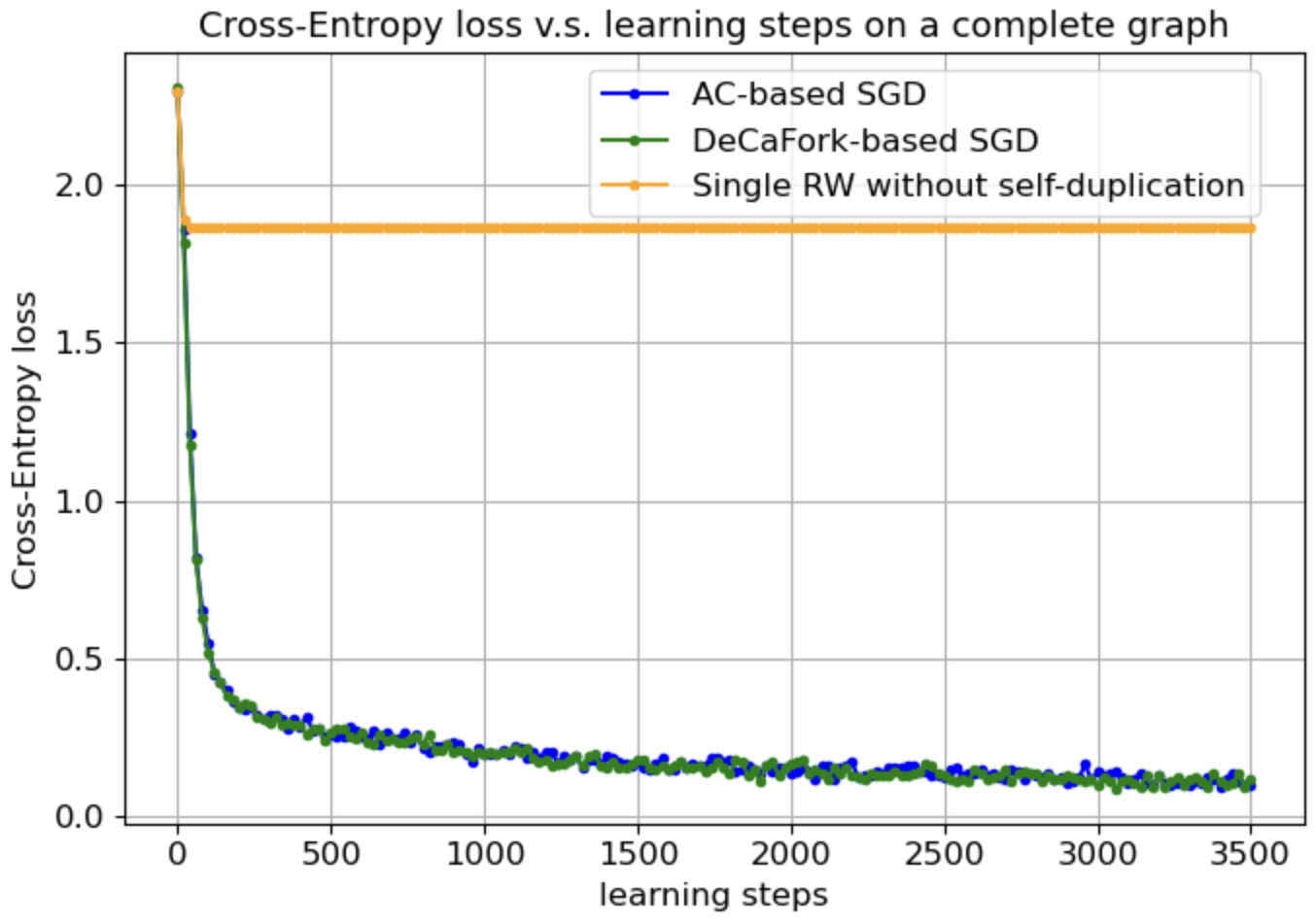}
    \caption*{(a) complete graph}
  \end{minipage}
  \hfill
  \begin{minipage}[b]{0.45\linewidth}
    \centering
    \includegraphics[width=\linewidth, height=0.9\linewidth]{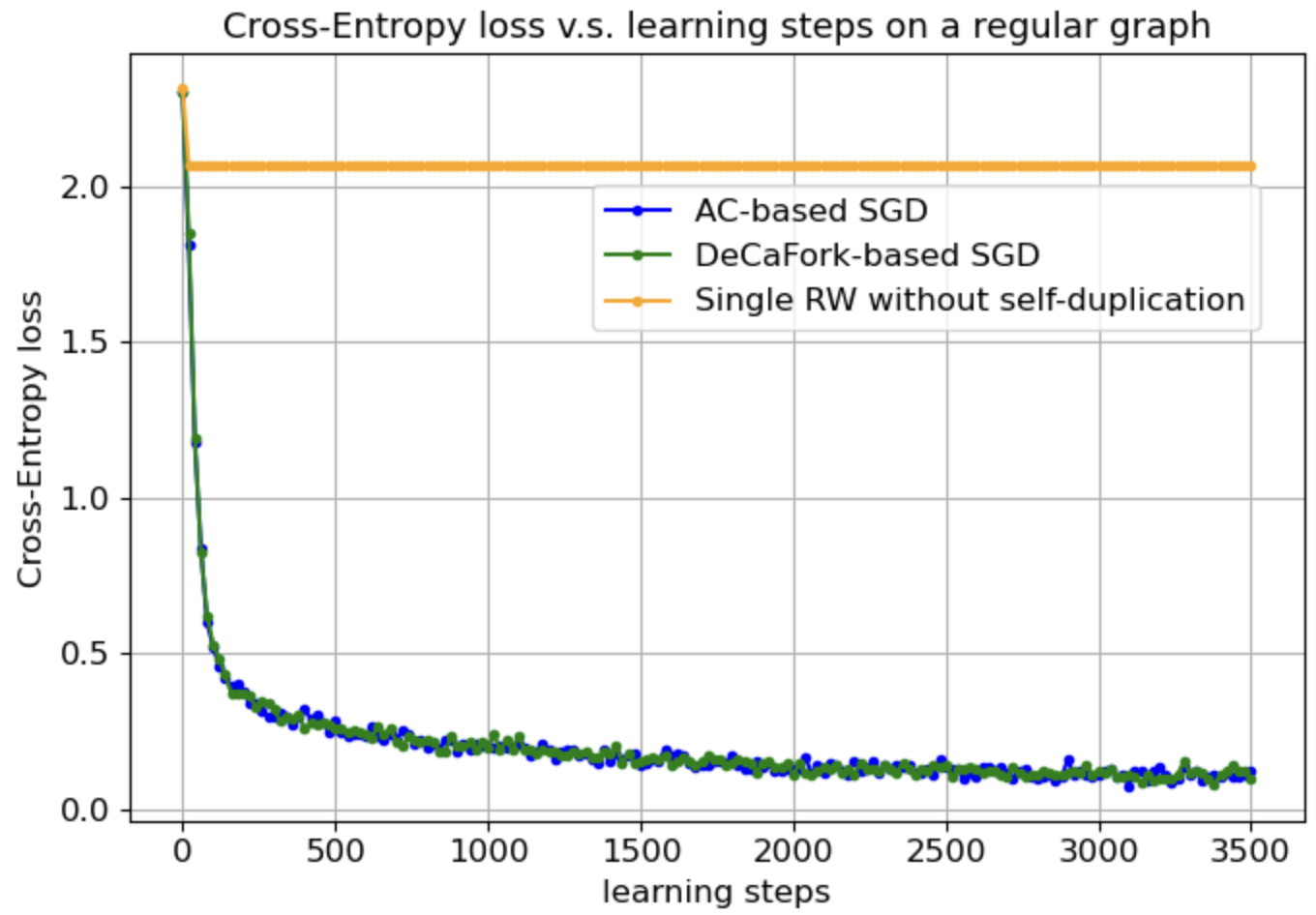}
    \caption*{(b) random regular graph}
  \end{minipage}
  \hfill
  \begin{minipage}[b]{0.45\linewidth}
    \centering
    \includegraphics[width=\linewidth, height=0.9\linewidth]{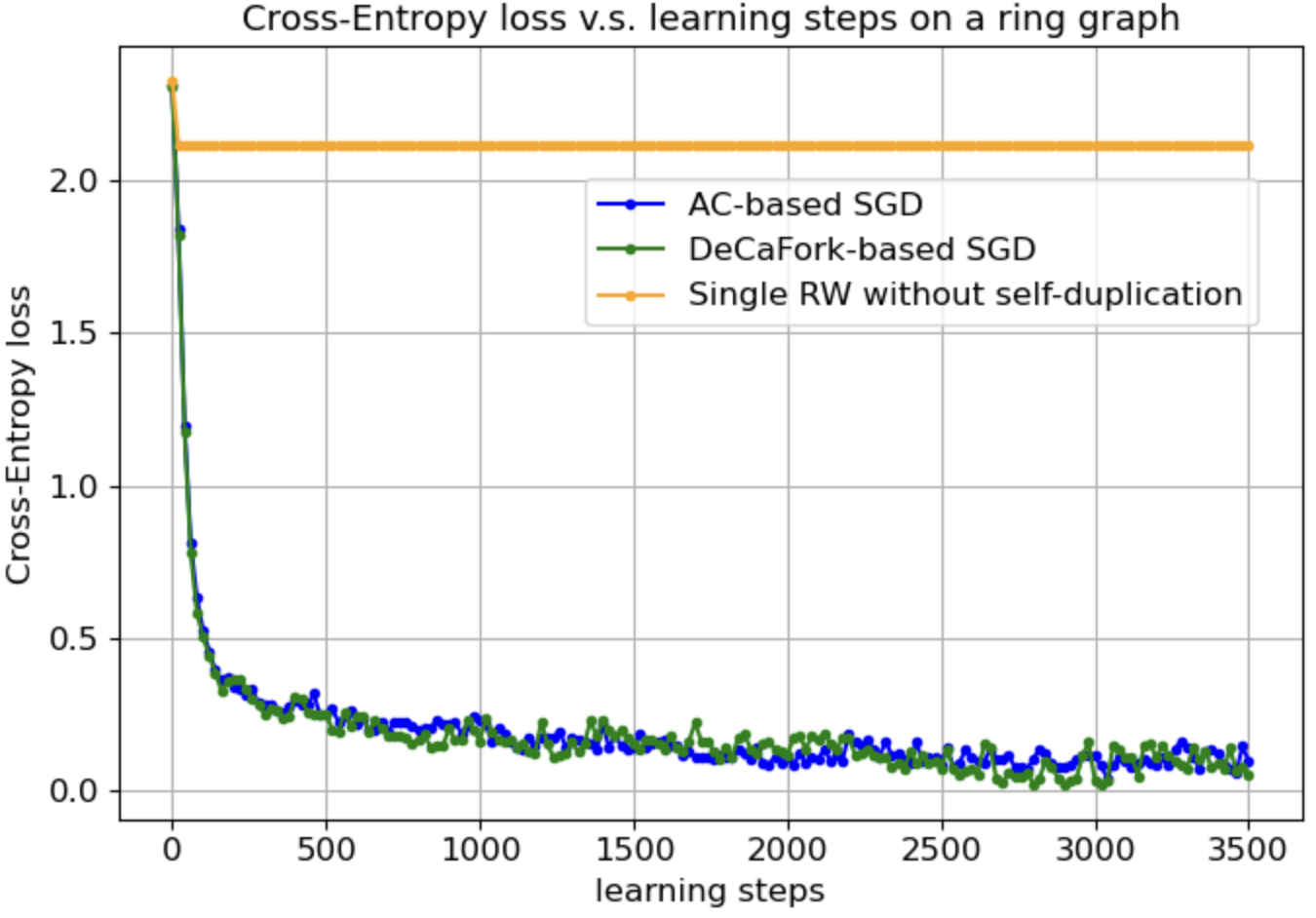}
    \caption*{(c) ring topology}
  \end{minipage}
  \hfill
  \begin{minipage}[b]{0.45\linewidth}
    \centering
    \includegraphics[width=\linewidth, height=0.9\linewidth]{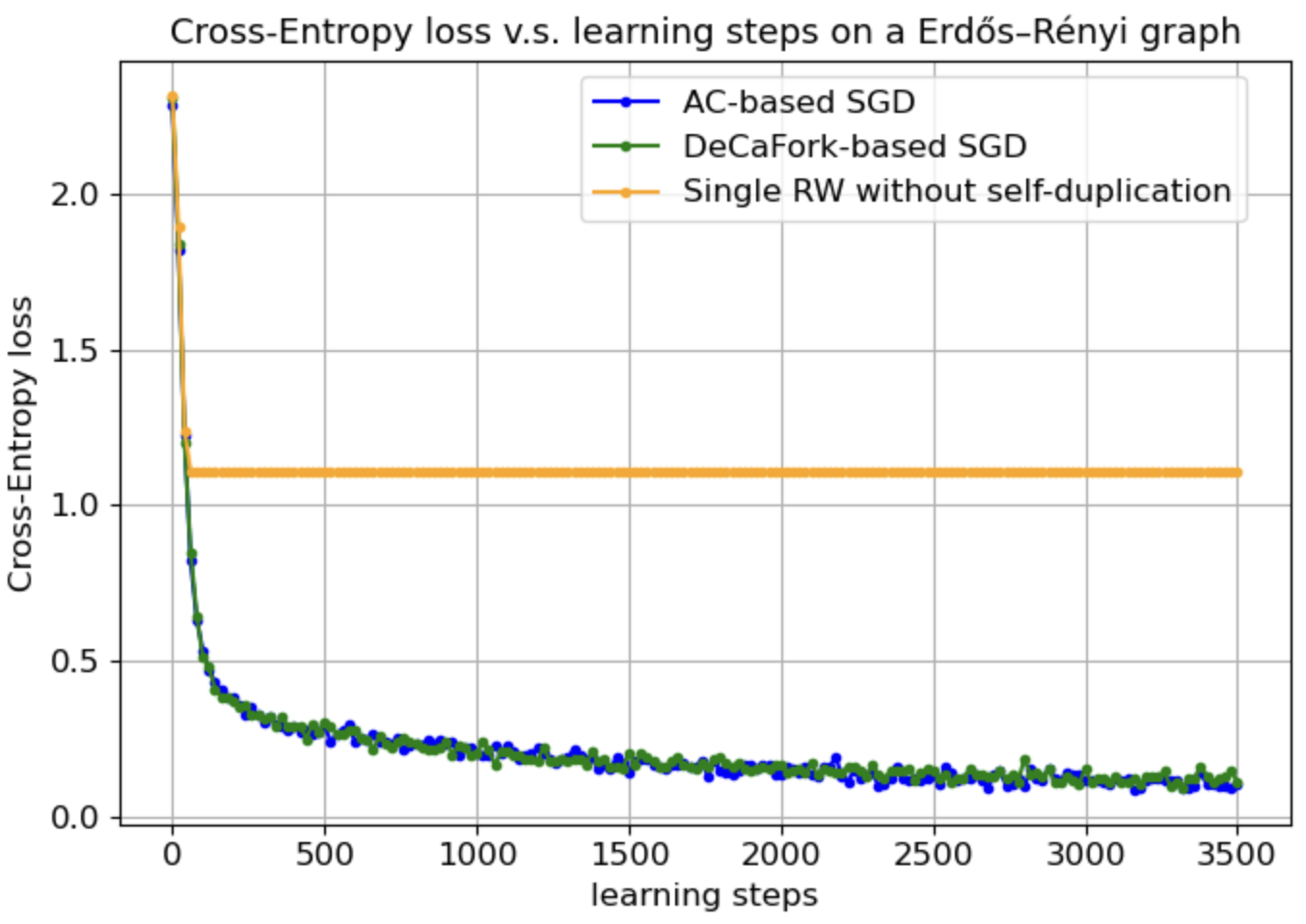}
    \caption*{(c) Erd\H{o}s--R\'enyi graph}
  \end{minipage}
\caption{Loss function v.s. learning steps on different graphs under \iid partitioning of the real-world dataset.}
\label{fig:Convergencerealiid}
\end{figure}

\begin{figure}[h]
\centering
  \centering
  \begin{minipage}[b]{0.45\linewidth}
    \centering
    \includegraphics[width=\linewidth, height=0.9\linewidth]{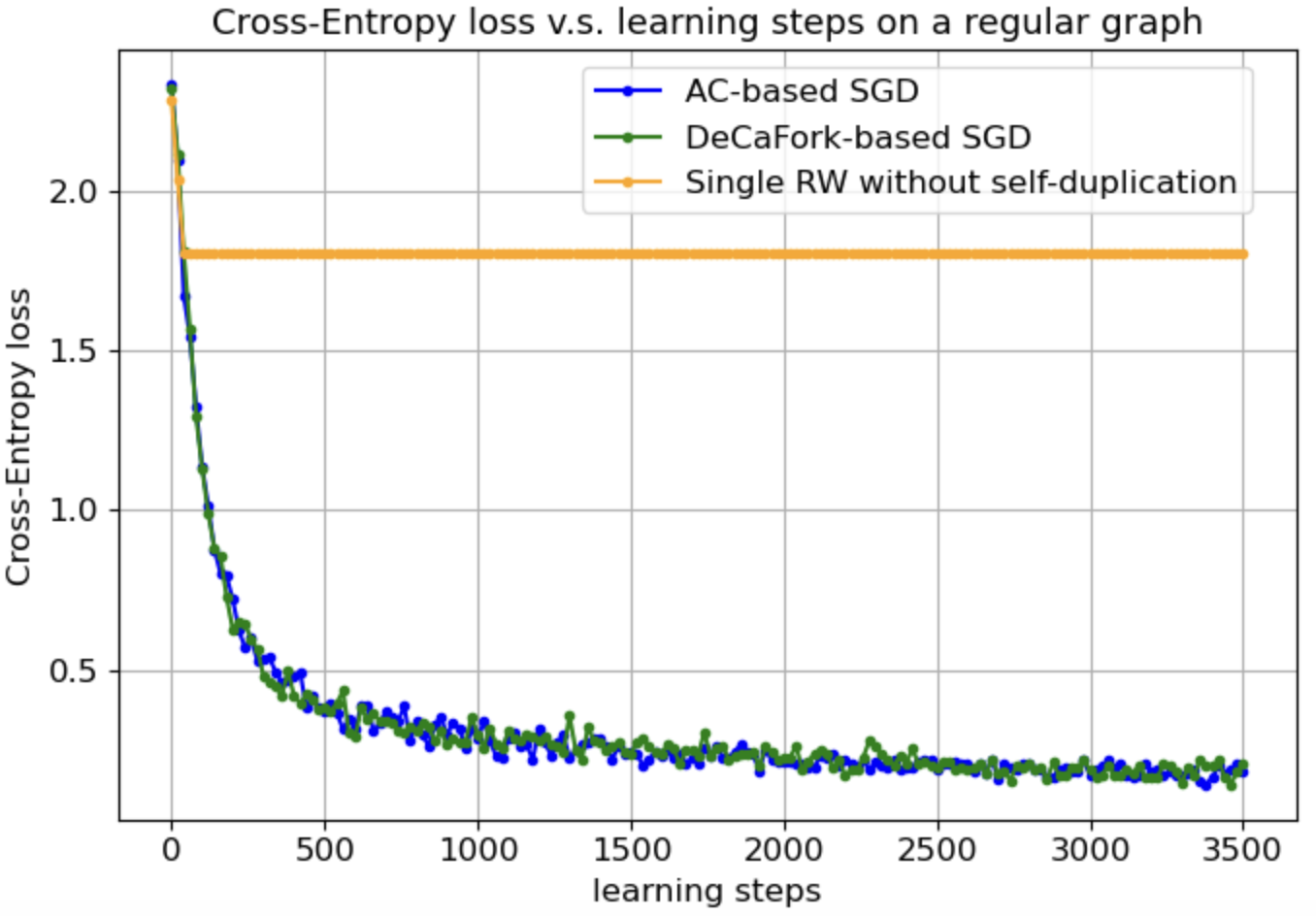}
    \caption*{(b) random regular graph}
  \end{minipage}
  \hfill
  \begin{minipage}[b]{0.45\linewidth}
    \centering
    \includegraphics[width=\linewidth, height=0.9\linewidth]{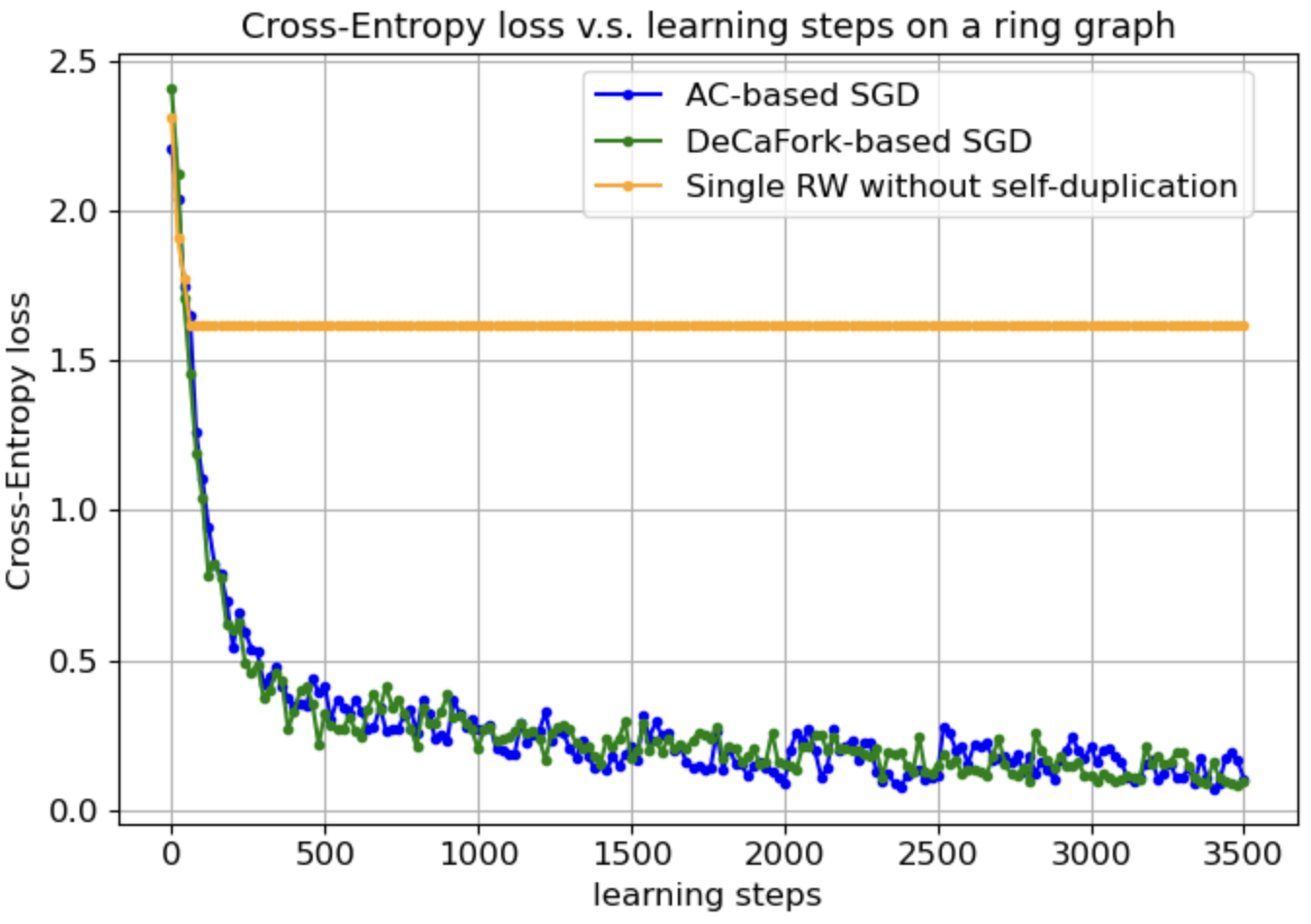}
    \caption*{(c) ring topology}
  \end{minipage}
  \hfill
  \begin{minipage}[b]{0.45\linewidth}
    \centering
    \includegraphics[width=\linewidth, height=0.9\linewidth]{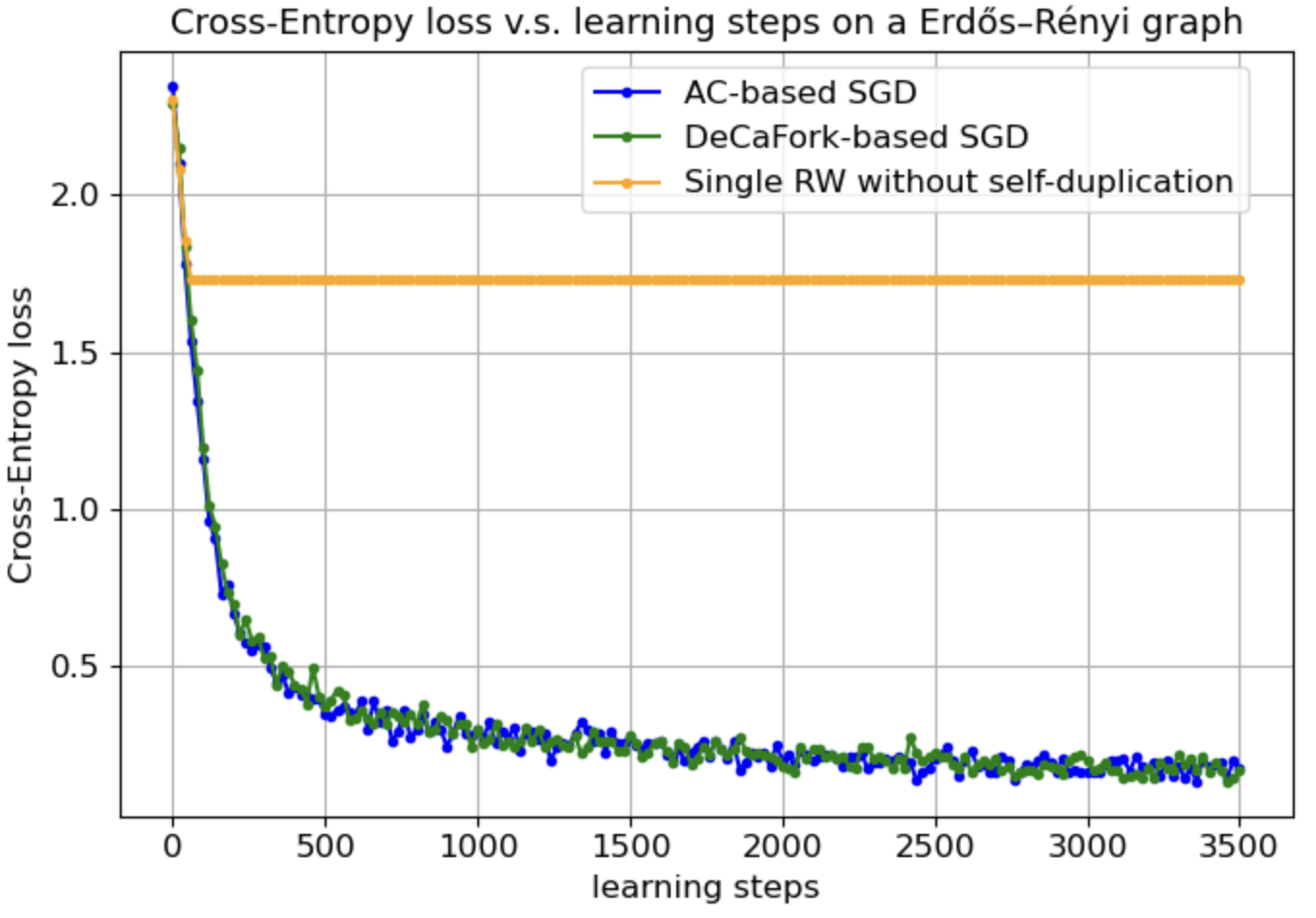}
    \caption*{(c) Erd\H{o}s--R\'enyi graph}
  \end{minipage}
\caption{Loss function v.s. learning steps on different graphs under non-\iid partitioning of the real-world dataset.}
\label{fig:ConvergencerealNoniid}
\end{figure}

Fig.~\ref{fig:Convergencerealiid} and Fig.~\ref{fig:ConvergencerealNoniid} show the convergence behavior of RW-SGD under the \ac and \DeCa algorithms, on the public benchmark dataset~\cite{deng2012mnist}, using \iid and non-\iid data partitioning across nodes, respectively. In Fig.~\ref{fig:Convergencerealiid} and Fig.~\ref{fig:ConvergencerealNoniid}, each subfigure plots the global loss (on the $y$-axis) over time steps (on the $x$-axis). In both partitioning settings, the global loss drops quickly to near-zero levels and remains stable thereafter, confirming that the algorithm converges. 
Moreover, RW-SGD exhibits almost identical convergence behavior under all three self-duplication mechanisms. This aligns with the observations in Fig.~\ref{fig:Convergence}, as all three algorithms share the same RW transition matrix, differing only in their replication mechanisms, not in how the RWs traverse the network.

\begin{table}[h]
\centering
\begin{tabular}{|c|c|c|c|c|}
\hline
& Complete & Regular & Ring & Erd\H{o}s--R\'enyi \\
\hline
\makecell{A single SGD,\\ no the Pac-Man} & $0.9749$ & $0.9748$ & $0.9743$ & $0.9751$\\
\hline
AC & $0.9732$ & $0.9779$ & $0.9764$ & $0.9709$ \\
\hline
DeCaFork & $0.9725$ & $0.9731$ & $0.9749$  & $0.9712$ \\
\hline
\makecell{A single SGD, \\ no self-duplication} & $0.7104$ & $0.6615$ & $0.5390$& $0.7885$\\
\hline
\end{tabular}
\caption{Testing accuracies on different graphs under \iid partitioning.}
\label{tab:Accuracyiid}
\end{table}

Finally, we evaluate the performance of the final models across different graphs using testing accuracies as the performance metric. We first examine the surrogate optimization problem~\eqref{eq:SurOpt}, particularly the quasi-stationary distribution $\nu^{(1)}$ (with $\zeta=1$).  Given the target distribution $\pi = \left(\frac{1}{100}, \dots, \frac{1}{100} \right)$, Theorem~\ref{thm:Effectiveness} implies that $\nu^{(1)}$ is uniform over the $99$ benign nodes in complete, random regular, and ring graphs, and approximately uniform in the Erd\H{o}s--R\'enyi graph due to its near-regular structure. This indicates that in the presence of a Pac-Man node, active RWs under \ac and \DeCa visit benign nodes uniformly or near-uniformly.

Table~\ref{tab:Accuracyiid} presents the performances of final models under \iid data partitioning. From this table, we observe that the performances are nearly identical across all graph types, regardless of whether a Pac-Man node is present. In the case of the single-SGD baseline without Pac-Man, the RW visits all nodes uniformly, and each node holds an \iid portion of the dataset, yielding a faithful representation of the full data. In the Pac-Man cases, although the Pac-Man node is never visited, the benign nodes share the same data distribution (due to the \iid partitioning) and are accessed uniformly (or approximately so) by active RWs. As the dataset is large ($100$ nodes with $600$ samples each), excluding one node's data has negligible effect, and the final models still perform well.

\begin{table}[h]
\centering
\begin{tabular}{|c|c|c|c|c|}
\hline
& Complete & Regular & Ring & Erd\H{o}s--R\'enyi \\
\hline
\makecell{A single SGD,\\ no Pac-Man} & $0.9764$ & $0.9719$ & $0.9782$ & $0.9751$\\
\hline
AC & $0.9613$ & $0.9611$ & $0.9617$ & $0.9596$ \\
\hline
DeCaFork & $0.9620$ & $0.9606$ & $0.9596$  & $0.9621$ \\
\hline
\makecell{A single SGD, \\ no self-duplication} & $0.5508$ & $0.5166$ & $0.5539$& $0.5812$\\
\hline
\end{tabular}
\caption{Testing accuracies on different graphs under non-\iid partitioning.}
\label{tab:Accuracymain}
\end{table}

Table~\ref{tab:Accuracymain} shows the results under non-\iid partitioning. We evaluate the performance of the final models across different graphs using testing accuracies as the performance metric. Table~\ref{tab:Accuracymain} shows that the presence of a Pac-Man node impacts performance: the performance in the single SGD without the Pac-Man outperforms those in the Pac-Man cases, as the RW has full access to all nodes and thus to the complete data distribution; in contrast, the active RWs in the Pac-Man cases never access the Pac-Man node, and the remaining benign nodes---now with heterogeneous data (due to the non-\iid partitioning)---no longer provide a representative sample of the entire dataset, resulting in degraded performance. 
\end{document}